\documentclass[pdflatex,sn-mathphys-num]{sn-jnl}% Math and Physical Sciences Numbered Reference Style
\usepackage{graphicx}%
\usepackage{multirow}%
\usepackage{amssymb}%
\usepackage{amsthm}%
\usepackage{mathrsfs}%
\usepackage[title]{appendix}%
\usepackage{xcolor}%
\usepackage{textcomp}%
\usepackage{manyfoot}%
\usepackage{booktabs}%
\usepackage{algorithm}%
\usepackage{algorithmicx}%
\usepackage{algpseudocode}%
\usepackage{listings}%
\usepackage{amsfonts} % For \mathbb
\usepackage{amsmath}  % For \text and equations
\usepackage{float}    % For [H] table placement
\usepackage{tabularx}
\usepackage{adjustbox} % for tables scaled inside threeparttable (sn-jnl redefines \table)
\usepackage{subcaption}
 
\usepackage{tikz}
\usetikzlibrary{arrows.meta, positioning, fit, backgrounds, shapes.geometric}
 
\theoremstyle{thmstyleone}%
\theoremstyle{thmstyletwo}%

\theoremstyle{thmstylethree}%

\newcommand{\sd}[1]{$\pm$#1}
\begin{document}
 
\title[Article Title]{Neuro-Symbolic Graph Transformers for Hierarchical Intention Recognition from Multimodal Human Activity Sequences}
 
 
\author[1]{\fnm{Farnaz} \sur{Soleimani}}\email{farnaz.soleimani@u-pec.fr}

\author[1]{\fnm{Abdelghani} \sur{Chibani}}\email{abdelghani.chibani@u-pec.fr}
 
\author[1]{\fnm{Yacine} \sur{Amirat}}\email{amirat@u-pec.fr}
 
\author*[1]{\fnm{Ghazaleh} \sur{Khodabandelou}}\email{ghazaleh.khodabandelou@u-pec.fr}
\affil*[1]{\orgdiv{LISSI Laboratory}, \orgname{University of Paris-Est Créteil (UPEC)}, \orgaddress{\street{{122 Rue Paul Armangot}}, \city{Vitry-sur-Seine}, \postcode{94400}, \country{France}}}

 
\abstract{Proactive human-robot collaboration requires autonomous
systems to infer not only immediate actions but also the hierarchical
intentions that govern human behaviour. Current deep learning
approaches focus predominantly on flat action classification and do
not capture the structural dependencies linking kinematic motion to
high-level goals. A Neuro-Symbolic Graph Transformer (NSGT) is
proposed, within which a Human Reasoning Graph (HR-G) is integrated
into the multimodal self-attention mechanism through learnable
topological biases. The architecture is augmented with a
differentiable First-Order Logic (FOL) loss that acts as a structural
regulariser, promoting ontological consistency across four hierarchy
levels. Evaluation is performed on the NTU-Hierarchy benchmark, a
four-level ontological extension of NTU RGB+D 120 spanning 58 atomic
actions, 39 composite activities, 28 low-level intentions, and 8
high-level intentions, under a subject-disjoint protocol with a
compositional generalisation split. Because the hierarchy labels and
the FOL constraints derive from the same ontology, hierarchy
consistency is treated as an objective-aligned diagnostic rather than
independent evidence of reasoning. A top-1 action accuracy of 87.32\%
is attained in the base configuration; adding the FOL loss raises the
hierarchy consistency rate monotonically with the logic weight to
78.79\% at $\lambda=0.5$, while 87.37\% action accuracy is retained,
so structural coherence is improved at no cost to accuracy. The full
model surpasses sequential, multilayer, and external relational-graph
baselines at the intention levels and attains the highest hierarchy
consistency and joint accuracy on the compositional split.
Interpretable evidence is provided by the learned HR-G weights, with
low-level-intention co-membership edges most strongly exploited.
}
 
%%================================%%
%% Sample for structured abstract %%
%%================================%%

 
 
 
\maketitle
%% ═══════════════════════════════════════════════════════════════════════════════
\section{Introduction}\label{intro}
 
Accessible robotics platforms, including cost-effective unmanned ground
vehicles developed for education and research
\cite{corpuz2025novel},
have broadened the settings in which proactive human-robot
collaboration is deployed, which motivates robust hierarchical
intention recognition on multimodal input.
 
%% ═══════════════════════════════════════════════════════════════════════════════
\subsection{Motivation and Context}
Human-centric autonomous systems, from surgical assistants and warehouse robots to social care agents, must do far more than recognize isolated physical gestures. To act proactively and safely, such systems must infer \emph{why} a person is performing a sequence of actions: their underlying goals, plans, and intentions. Human Intention Recognition (HIR), particularly long-term, hierarchical intention recognition, remains a formidable challenge. Unlike low-level action recognition, which maps raw sensory inputs to immediate kinematic behaviors, intention recognition requires modeling complex, multi-granular spatiotemporal dependencies~\cite{kekana2025review}. These dependencies span a hierarchical continuum: from atomic actions, to composite activities, to low-level intentions, and ultimately to high-level intentions. Interpreting these hierarchical structures from multimodal data streams (e.g., RGB, Infrared, Depth, and Skeletal sequences) is paramount for proactive and safe decision-making in autonomous systems, as evidenced by the growing deployment of intention-aware systems in collaborative manufacturing, healthcare assistance, and autonomous navigation \cite{zhao2025multimodal, kekana2025review}.
Recent advancements in deep representation learning, notably Graph Neural Networks (GNNs) and Vision Transformers (ViTs), have demonstrated remarkable efficacy in modeling non-Euclidean skeletal kinematics and high-dimensional multimodal streams~\cite{plizzari2021spatial}. However, relying on purely data-driven, black-box architectures presents significant theoretical and practical limitations. Standard spatiotemporal transformers operate on the premise of statistical correlation, optimizing a loss function over large-scale distributions without an intrinsic comprehension of causal or structural realism. Consequently, these models are fundamentally prone to generating logically inconsistent predictions across the intention hierarchy. For instance, a purely connectionist model might classify a subject's atomic action as \emph{drinking water} ($a \in \mathcal{A}$) while paradoxically predicting a high-level intention such as \emph{hostile social interaction} ($I_{high} \in \mathcal{I}$), a pairing that no path in the intention hierarchy licenses, simply due to dataset bias or transient sensor noise. This lack of robust logical constraints severely restricts their reliability in autonomous applications~\cite{rudin2019stop}.
 
To transcend the fragility of purely statistical pattern recognition, structural
priors must be enforced to ground the network's predictions in logical reality.
Neuro-symbolic artificial intelligence has recently emerged as a principled paradigm to bridge the gap between neural perception and symbolic reasoning.
 
By augmenting Graph Transformers with declarative First-Order Logic (FOL)
constraints, the prerequisite, temporal, and hierarchical axioms that govern
human behavior can be explicitly encoded.
Let $\mathcal{R}$ denote a set of FOL rules parameterized by a continuous truth
domain.
By embedding these logical predicates into the latent representations of a
Neuro-Symbolic Graph Transformer (NSGT), the optimization objective is extended
to encompass both data fidelity and logical satisfiability:
\begin{equation}
    \mathcal{L}_{\mathrm{total}}
    = \mathcal{L}_{\mathrm{CE}}(\mathcal{X}, \mathcal{Y})
    + \lambda\,\mathcal{L}_{\mathrm{FOL}}(\mathcal{R})
    \label{eq:total_intro}
\end{equation}
This integration promotes hierarchical consistency, penalizing
inferred high-level intentions that are not licensed by their constituent
low-level activities and actions, yielding a system that
is both highly expressive and structurally sound.
The concrete instantiation of $\mathcal{L}_{\mathrm{CE}}$ and
$\mathcal{L}_{\mathrm{FOL}}$ is provided in Section~\ref{sec:fol_loss}.
 
\subsection{Problem Statement and Contributions}
Despite the critical imperative for hierarchical intention recognition in autonomous systems, the advancement of this field is severely constrained by two fundamental bottlenecks.
 
\textbf{The Data Bottleneck:} A profound scarcity of large-scale, multimodal datasets featuring deep, hierarchical intention annotations is observed. Existing multimodal benchmarks, such as NTU RGB+D~\cite{liu2019ntu}, predominantly provide flat, atomic action labels. Conversely, datasets that include rich hierarchical task annotations, such as the Breakfast~\cite{kuehne2014language} or Charades~\cite{sigurdsson2016hollywood} datasets, are largely unimodal, limited in scale, or restricted to narrow domains. Consequently, evaluating complex, multi-level intention reasoning over comprehensive multimodal streams remains difficult using standard publicly available datasets. Since no suitable benchmark existed for this task, one was constructed, a process that constitutes a core contribution of this work.
 
\textbf{The Modeling Bottleneck:} When tasked with predicting multiple levels of cognitive abstraction, standard deep neural networks often formulate the objective as a set of parallel classification tasks. Without an explicit structural mechanism, such models may produce ontologically inconsistent predictions across levels. This motivates evaluating whether graph priors and soft logical constraints improve cross-level coherence.
 
To resolve these limitations, this paper introduces a holistic neuro-symbolic framework that bridges the gap between deep multimodal perception and structured logical reasoning. The core contributions of this work are summarized as follows:
 
\begin{itemize}
    \item \textbf{NTU-Hierarchy Dataset:} A novel, hierarchically labeled dataset is synthesized based on the multimodal NTU RGB+D corpus. By mapping atomic labels to a four-level ontological extension ($\text{Action} \rightarrow \text{Activity} \rightarrow \text{LLI} \rightarrow \text{HLI}$), which explicitly encodes the contextual and temporal relationships between cognitive levels, a large-scale multimodal benchmark is established. This dataset is specifically designed for hierarchical intention reasoning. The benchmark is released with a subject-disjoint evaluation protocol and a dedicated compositional generalization split, so that transfer to unseen performers and to unseen intention-level associations is measured explicitly.
 
    \item \textbf{Neuro-Symbolic Graph Transformer (NSGT):} A structural Human Reasoning Graph (HR-G) is integrated into Transformer self-attention through learnable transition and ontological biases ($\beta$). The model learns the relative contribution of the three edge types. Under the reported experiments, the low-level-intention co-membership coefficient ($\beta_{\mathrm{lli}}$) becomes the largest, while temporal-transition and same-activity coefficients remain smaller and positive.
 
    \item \textbf{Differentiable First-Order Logic (FOL) Loss:} Declarative hierarchy rules are translated into soft penalties compatible with gradient-based optimization. The resulting regularizer, controlled by $\lambda$, penalizes inconsistent cross-level predictions. In the reported experiments it primarily improves the objective-aligned hierarchy-consistency diagnostic, while per-level classification metrics change only slightly.
\end{itemize}
 
The proposed framework aligns with the objectives of applied artificial intelligence by bridging perception and reasoning in complex multimodal environments, with potential applications in assistive robotics, smart environments, and human-centered AI systems.
 
%%═══════════════════════════════════════════════════════════════════════════════
\section{Related Work}
\label{sec:related}
 
Complementary directions in human activity recognition include
ensemble classifiers for smartphone-sensor action data
\cite{raja2025hstcn}
and federated learning for sensor-based recognition, which addresses
privacy and distributed data
\cite{wu2025review}.
Within robotics, neural control methods such as dynamic gain control
for humanoid trajectory tracking
\cite{Trujillo2025EnhancingTT}
illustrate the perception-to-control pipeline that intention
recognition is intended to inform.
 
The idea of relaxing symbolic constraints into differentiable losses
has precedents, including semantic loss functions for deep learning
with symbolic knowledge
\cite{xu2018semantic}
and Logic Tensor Networks \cite{serafini2016logic, badreddine2022logic}. The contribution
here is not the differentiable-logic idea in isolation, but its
integration with a learnable graph-structured attention bias over an
ontological hierarchy, applied jointly to multi-level intention
recognition. To the best of current knowledge, the combination of an
ontology-derived attention bias whose edge-type weights are learned
end-to-end with a hierarchical differentiable-logic regulariser, within
a single multimodal transformer, has not been previously reported.
 
%% ═══════════════════════════════════════════════════════════════════════════════
\subsection{Multimodal Action and Activity Recognition}
The paradigm of human behavior analysis has fundamentally shifted from unimodal representation learning to the fusion of heterogeneous sensor streams. Early methodologies predominantly relied on isolated RGB sequences, rendering them highly susceptible to visual occlusions, illumination variations, and background clutter \cite{simonyan2014two}. While large-scale RGB benchmarks such as Kinetics-700 provide powerful pre-training foundations, their unimodal nature limits their utility for robust real-world deployment \cite{smaira2020short}. To mitigate these vulnerabilities, contemporary architectures explicitly fuse multimodal data, incorporating RGB, skeletal kinematics, infrared (IR), and depth maps, to extract robust, environment-invariant spatiotemporal features \cite{baltruvsaitis2018multimodal, liu2019ntu}. Notably, the integration of active sensors (IR and depth) with passive RGB streams has proven particularly effective: active modalities provide geometric and thermal cues that are invariant to ambient illumination, substantially stabilizing recognition performance under real-world lighting variability. Datasets such as MMAct \cite{kong2019mmact} further demonstrate the value of heterogeneous sensor fusion, integrating RGB, keypoints, accelerometers, gyroscopes, and even Wi-Fi signals, enabling cross-modal transfer and robustness against occlusion and appearance variation.
 
Graph-based architectures have emerged as the dominant paradigm for skeleton-based analysis, with Spatial Temporal Graph Convolutional Networks (ST-GCN) establishing the foundational framework for modeling non-Euclidean joint dependencies \cite{yan2018spatial, shi2019two}. While the integration of these diverse streams has substantially elevated state-of-the-art accuracy, a critical limitation persists in the literature: the vast majority of these models are architecturally constrained to classifying short-term, atomic actions. They optimize exclusively for immediate kinematic feature mapping, failing to execute the long-term temporal abstraction fundamentally required to deduce overarching human goals.
 
\subsection{Hierarchical Activity and Intention Understanding}
Transitioning from basic kinematic action recognition to the comprehension of complex activities and latent intentions necessitates modeling temporal dynamics over significantly extended horizons. Temporal modeling approaches, including Temporal Convolutional Networks (TCNs) and sequential segmentation models, have demonstrated strong performance on activity-level recognition \cite{farha2019ms}. Previous attempts to bridge low-level sensory inputs with higher-level semantic goals have primarily relied on sequential networks or probabilistic graphical models \cite{kekana2025review}. However, these approaches typically discretize the reasoning process into disjointed stages or limit their scope to a shallow, two-tier hierarchy (e.g., mapping actions directly to activities). While some works provide coarse two-tier annotations, such as the Breakfast dataset \cite{kuehne2014language}, the joint modeling of a deep, four-level semantic continuum directly from continuous multimodal streams remains unaddressed.
The broader intention recognition landscape includes specialized benchmarks for pedestrian behavior (JAAD \cite{kotseruba2016joint}, PIE \cite{rasouli2019pie}) and industrial assembly (MCV-Intention \cite{ma2025mcv}). While these datasets introduce contextual and goal-directed annotations, they remain domain-specific, constrained to traffic scenarios or factory floors, and none provides the deep, four-level semantic hierarchy ($\text{Action} \rightarrow \text{Activity} \rightarrow \text{LLI} \rightarrow \text{HLI}$) required for general-purpose cognitive reasoning over heterogeneous multimodal streams.
Consequently, there remains a pronounced gap in the current literature: very few works attempt to jointly model a deep, multi-granular semantic hierarchy. Specifically, the holistic inference of an ontology spanning $\text{Action} \rightarrow \text{Activity} \rightarrow \text{LLI} \rightarrow \text{HLI}$ directly from continuous, multimodal video streams remains largely unaddressed, primarily due to the severe computational bottlenecks and contextual ambiguity inherent in long-term video understanding.
 
\subsection{Neuro-Symbolic AI in Sequence Modeling}
To address the fragility and logical inconsistencies of purely connectionist deep learning architectures, Neuro-Symbolic AI has emerged as a rigorous paradigm. This approach seeks to amalgamate the scalable pattern recognition of neural networks with the structural cognitive soundness of declarative logic \cite{hitzler2022neuro, garcez2023neurosymbolic}. Differentiable logic frameworks, such as DeepProbLog \cite{manhaeve2018deepproblog} and Logic Tensor Networks \cite{badreddine2022logic}, have demonstrated that declarative logical constraints can be embedded as differentiable loss terms, enabling end-to-end gradient-based optimization while preserving symbolic soundness. This trend is further evidenced by recent benchmarks such as SHREC \cite{lee2026social}, which evaluates foundational models on social reasoning tasks including intention tracking, engagement monitoring, and norm compliance, revealing a persistent gap between machine and human performance in non-verbal reasoning channels.
 
While recent neuro-symbolic methods have demonstrated considerable efficacy, their successful application has been predominantly restricted to static Visual Question Answering (VQA) or computationally simple spatial reasoning tasks \cite{yi2019clevrer}. Their extension into the domain of complex, spatiotemporal \emph{heterogeneous sensor streams}, encompassing RGB, depth, IR, and skeletal sequences simultaneously, remains critically underexplored. This exact gap is bridged in the present work by embedding Differentiable First-Order Logic (FOL) directly within a Graph Transformer framework operating on multimodal sequential data. By functioning as a dynamic structural regularizer, the FOL loss penalizes hierarchically inconsistent predictions across continuous temporal sequences, so that predicted high-level intentions are encouraged to remain anchored to their constituent low-level actions.

%% ═══════════════════════════════════════════════════════════════════════════════
\section{Methodology}
\label{sec:methodology}
 
\begin{figure}[t]
  \centering
  \includegraphics[width=\linewidth]{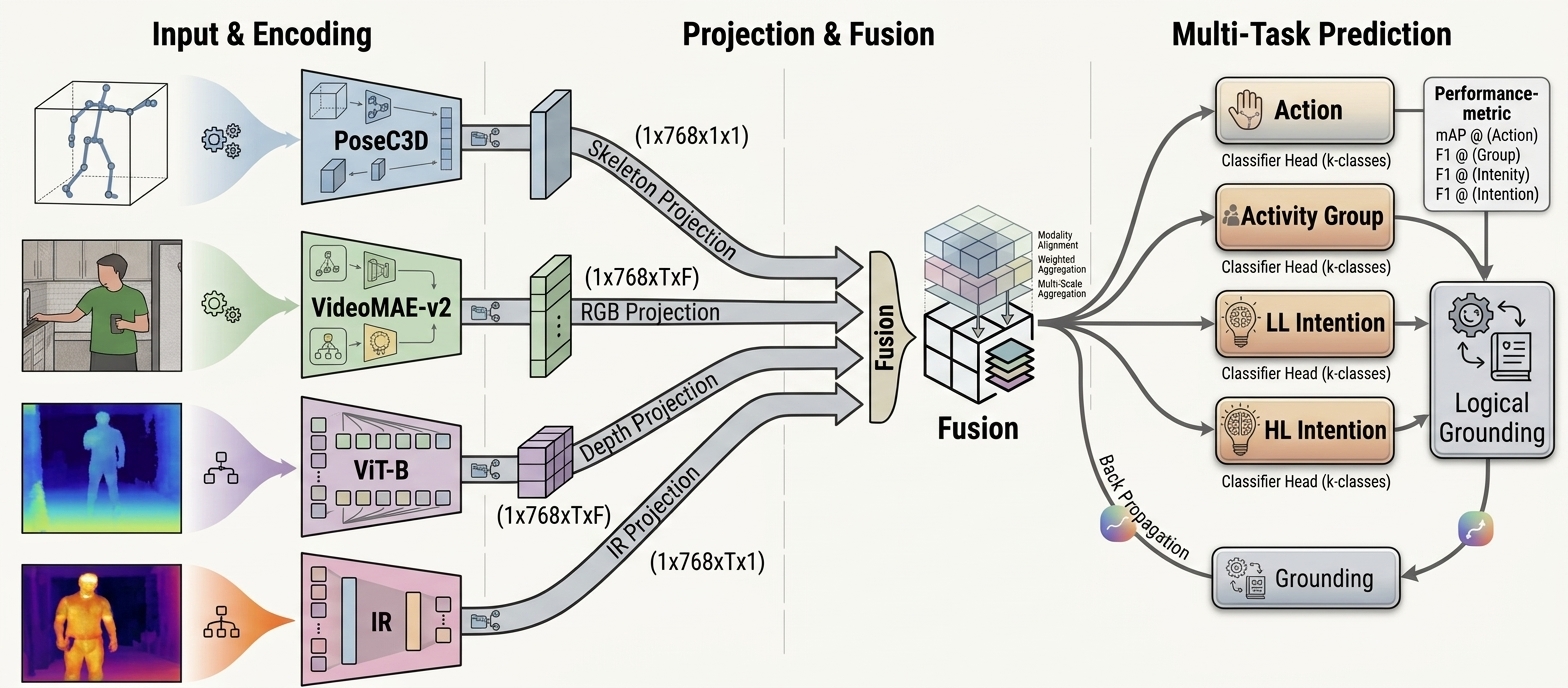}
  \caption{
    Overview of the NSGT framework.
    Four sensor streams (skeleton $\mathcal{X}_s$, RGB $\mathcal{X}_r$, infrared
    $\mathcal{X}_{ir}$, depth $\mathcal{X}_d$) are encoded by frozen pretrained encoders
    and concatenated into a single fused representation.
    The NSGT reasoning module augments standard self-attention
    with a learnable HR-G structural bias (Section~\ref{sec:nsgt}).
    Four prediction heads produce joint estimates of
    atomic actions, composite activities, Low-Level Intentions
    (LLI), and High-Level Intentions (HLI).
    The total loss combines cross-entropy with a differentiable
    FOL regulariser $\lambda_{\mathrm{FOL}} \times
    \mathcal{L}_{\mathrm{FOL}}$ that penalises ontological inconsistency
    across prediction levels.
  }
  \label{fig:pipeline}
\end{figure}
 
To address the challenges of multi-granular intention recognition,
a Neuro-Symbolic Graph Transformer (NSGT) is proposed.
The framework consists of a temporal encoder, a graph-structured
attention mechanism guided by a Human Reasoning Graph (HR-G),
and a differentiable First-Order Logic (FOL) regulariser.
Figure~\ref{fig:pipeline} provides an overview of the complete
pipeline.
 
%% ─────────────────────────────────────────────────────────────────
\subsection{Problem Formulation}
\label{sec:problem_formulation}
%% ─────────────────────────────────────────────────────────────────
 
Let $\mathcal{X} \in \mathbb{R}^{T \times d_{\mathrm{in}}}$ denote
an input sequence of multimodal video features over $T$ time steps,
where $d_{\mathrm{in}}$ is the dimensionality of the fused sensor
representation.
Hierarchical intention recognition is formulated as a joint
multi-task learning problem.
Given $\mathcal{X}$, the objective is to simultaneously predict:
 
\begin{enumerate}
  \item A sequence of atomic actions
    $\mathcal{A} = \{a_1, \dots, a_T\}$,
    where $a_t \in \mathbb{R}^{|\mathcal{V}_A|}$.
  \item A composite activity
    $\mathcal{C} \in \mathbb{R}^{|\mathcal{V}_C|}$.
  \item A Low-Level Intention (LLI)
    $\mathcal{L} \in \mathbb{R}^{|\mathcal{V}_L|}$.
  \item A High-Level Intention (HLI)
    $\mathcal{H} \in \mathbb{R}^{|\mathcal{V}_H|}$.
\end{enumerate}
 
Here $|\mathcal{V}_A|$, $|\mathcal{V}_C|$, $|\mathcal{V}_L|$,
and $|\mathcal{V}_H|$ denote the number of classes at each
ontological level.
The joint prediction is subject to ontological consistency
constraints $\Phi$ derived from the HR-G (Section~\ref{sec:hrg_logic}).
 
%% ─────────────────────────────────────────────────────────────────
\subsection{Hierarchical Transformer Backbone}
\label{sec:ht_backbone}
%% ─────────────────────────────────────────────────────────────────
 
A standard Transformer encoder \cite{vaswani2017attention} serves
as the temporal backbone.
The input sequence $\mathcal{X}$ is projected into a latent
dimension $d_{\mathrm{model}}$ via a linear layer followed by
layer normalisation, and sinusoidal positional encodings are added.
The sequence is then processed through $L$ layers of Multi-Head
Self-Attention (MHSA), producing per-step representations
$\{z_1^{(L)}, \dots, z_T^{(L)}\}$.
 
Four prediction heads share the Transformer representation and are chained through continuous representations rather than through decisions. The action head operates on the per-step representations. The pooled episode representation feeds the activity head directly and also feeds an MLP that produces the LLI hidden representation, which in turn feeds the LLI head and, through a further MLP, the HLI head. The two intention levels are therefore chained through continuous projections while the activity logits branch directly off the pooled representation, and no discrete activity or LLI prediction is passed to a later classification head.
 
Per-step action logits are computed directly from the sequence
representations:
\begin{equation}
  \hat{a}_t = W_{\mathrm{act}}\, z_t^{(L)}
  \;\in\; \mathbb{R}^{|\mathcal{V}_A|}
  \label{eq:action_head}
\end{equation}
 
A masked temporal mean-pool produces a single episode-level
embedding, ignoring padding positions:
\begin{equation}
  z_{\mathrm{seq}} =
  \frac{\sum_{t=1}^{T} m_t\, z_t^{(L)}}{\sum_{t=1}^{T} m_t},
  \quad m_t = \mathbb{1}[\text{clip } t \text{ is not padding}]
  \label{eq:pooling}
\end{equation}
 
The activity logits and the LLI intermediate representation
$z_{\mathrm{lli}}$ are then computed from $z_{\mathrm{seq}}$:
\begin{equation}
  \hat{c} = W_c\, z_{\mathrm{seq}}
  \;\in\; \mathbb{R}^{|\mathcal{V}_C|}
  \label{eq:activity_head}
\end{equation}
\begin{equation}
  z_{\mathrm{lli}} =
  \mathrm{MLP}_{\mathrm{act \to lli}}(z_{\mathrm{seq}})
  \label{eq:lli_repr}
\end{equation}
 
LLI logits are projected from $z_{\mathrm{lli}}$, which is
simultaneously forwarded to produce the HLI logits:
\begin{equation}
  \hat{l} = W_{\mathrm{lli}}\, z_{\mathrm{lli}}
  \;\in\; \mathbb{R}^{|\mathcal{V}_L|}
  \label{eq:lli_head}
\end{equation}
\begin{equation}
  \hat{h} = W_{\mathrm{hli}}\,
  \mathrm{MLP}_{\mathrm{lli \to hli}}(z_{\mathrm{lli}})
  \;\in\; \mathbb{R}^{|\mathcal{V}_H|}
  \label{eq:hli_head}
\end{equation}
 
This backbone, referred to as the Hierarchical Transformer (HT),
is used as a controlled baseline in the ablation study
(Section~\ref{sec:ablation_joint}) to isolate the contribution
of graph-structured attention.
 
%% ─────────────────────────────────────────────────────────────────
\subsection{Neuro-Symbolic Graph Transformer (NSGT)}
\label{sec:nsgt}
%% ─────────────────────────────────────────────────────────────────
 
The integration of symbolic ontological knowledge directly into
the neural attention mechanism is the core contribution of this
work.
In standard MHSA, every clip attends to every other clip
uniformly, with no knowledge of the semantic relationships between
the corresponding actions.
In the proposed NSGT, the HR-G is encoded as a learnable additive
bias on the attention logits, so that the graph structure actively
steers information routing during inference.
 
While prior graph transformer architectures have applied
structural biases to \emph{spatial} joint connectivity graphs
\cite{plizzari2021spatial}, the approach here applies them to
\emph{semantic and ontological} relationships between 58 action
classes and their hierarchical goals, which is a distinct
contribution.
 
\subsubsection*{HR-G Attention Bias}
 
An additive bias matrix $B \in \mathbb{R}^{T \times T}$ is
computed once per batch from the predicted action identities of the input clips
(Section~\ref{sec:hrg_logic}), using pre-built lookup tables derived from the HR-G.
The same bias is then passed to all $L$ Transformer layers.
The elements of $B$ are defined as a weighted sum of three
topological relations:
\begin{equation}
  B_{ij} =
  \beta_{\mathrm{trans}}\,
    \mathbb{1}\bigl[\hat a_j \in \mathcal{N}_{\mathrm{trans}}(\hat a_i)\bigr]
  + \beta_{\mathrm{act}}\,
    \mathbb{1}\bigl[\mathcal{A}(\hat a_i) \cap \mathcal{A}(\hat a_j)
    \neq \emptyset\bigr]
  + \beta_{\mathrm{lli}}\,
    \mathbb{1}\bigl[\mathcal{L}(\hat a_i) \cap \mathcal{L}(\hat a_j)
    \neq \emptyset\bigr]
  \label{eq:hrg_bias}
\end{equation}
 
where:
\begin{itemize}
  \item $\mathcal{N}_{\mathrm{trans}}(\hat a_i)$ is the set of valid
    successor actions of the predicted action $\hat a_i$ in the semantic transition graph
    (Level~1 edges).
  \item $\mathcal{A}(\hat a_i)$ is the set of activities that the
    predicted action $\hat a_i$ belongs to (Level~2 edges).
  \item $\mathcal{L}(\hat a_i)$ is the set of LLIs reachable from
    $\hat a_i$ via the ontology (Level~3 edges).
  \item $\beta_{\mathrm{trans}},\, \beta_{\mathrm{act}},\,
    \beta_{\mathrm{lli}} \in \mathbb{R}$ are scalar parameters
    learned end-to-end via backpropagation, initialised to small
    positive values.
\end{itemize}
 
Each bias scalar $\beta_{\mathrm{trans}}$, $\beta_{\mathrm{act}}$ and $\beta_{\mathrm{lli}}$ is clamped to $[-3, 3]$ before the weighted sum, which bounds every entry of $B$ and prevents the graph structure from dominating the attention logits during early training.
The matrix is then expanded across all attention heads and added
to the scaled dot-product attention scores:
\begin{equation}
  \mathrm{Attention}(Q, K, V) =
  \mathrm{Softmax}\!\left(
    \frac{QK^\top}{\sqrt{d_{\mathrm{model}}}} + B
  \right) V
  \label{eq:nsgt_attn}
\end{equation}
 
By treating $\beta_{\mathrm{trans}}$, $\beta_{\mathrm{act}}$,
and $\beta_{\mathrm{lli}}$ as learnable parameters rather than
fixed weights, the network adapts the contribution of the three
predefined edge types; it does not learn or alter the graph topology.
The converged values reported in Section~\ref{sec:graph_impact}
provide interpretable evidence of which edge types the model
finds informative.
 
%% ─────────────────────────────────────────────────────────────────
\subsection{Differentiable First-Order Logic Loss}
\label{sec:fol_loss}
%% ─────────────────────────────────────────────────────────────────
 
When hierarchical prediction heads are trained independently,
the model can produce ontologically contradictory outputs,
for example predicting an activity that is incompatible with
the predicted LLI.
To address this, hierarchical dependencies are formalised as
First-Order Logic (FOL) axioms and relaxed into differentiable
penalty functions \cite{serafini2016logic}, so that logical
violations produce gradient signals during training.
 
Three types of rule are defined, covering 487 active constraints in total
(39 Type-A, 28 Type-B, 420 Type-C):
 
\textbf{Type A, activity to LLI implication (39 rules):}
\begin{equation}
  \mathrm{activity}(X, A) \;\to\;
  \bigvee_{L \in \mathcal{V}(A)} \mathrm{lli}(X, L)
  \label{eq:typeA}
\end{equation}
where $\mathcal{V}(A)$ is the set of valid LLIs for activity $A$.
 
\textbf{Type B, LLI to HLI implication (28 rules):}
\begin{equation}
  \mathrm{lli}(X, L) \;\to\;
  \bigvee_{H \in \mathcal{V}(L)} \mathrm{hli}(X, H)
  \label{eq:typeB}
\end{equation}
 
\textbf{Type C, mutex constraints (420 rules):}
\begin{equation}
  \neg\bigl(\mathrm{activity}(X, A_1) \wedge
  \mathrm{activity}(X, A_2)\bigr)
  \quad \text{if } \mathcal{H}(A_1) \cap \mathcal{H}(A_2) =
  \emptyset
  \label{eq:typeC}
\end{equation}
where $\mathcal{H}(A)$ denotes the set of HLIs reachable from $A$.
As an example, predicting both \texttt{EatFood} and \texttt{Confront}
as concurrent activities in a single episode violates a Type-C rule,
because no high-level intention is reachable from both activities, so
they cannot logically co-occur.
 
The discrete rules are converted into continuous soft violations by
operating on probability vectors $p_A$, $p_L$, and $p_H$ derived from
the model outputs for activities, LLIs, and HLIs respectively. At the
activity and low-level intention levels, which admit more than one
valid label within an episode, per-class sigmoid marginals are used;
at the high-level intention level, which is single-label, a softmax
distribution is used.
This relaxation is necessary because hard IF/THEN thresholds
produce zero gradients and are invisible to backpropagation.
Soft probabilities instead produce continuous, non-zero gradient
signals that allow the model to correct ontological
contradictions iteratively during training.
 
For implication rules (Types A and B), the violation $v_\Rightarrow$
measures how much the antecedent probability exceeds the maximum
probability among its valid consequents:
\begin{equation}
  v_\Rightarrow(A, \mathcal{V}) =
  \max\!\left(0,\; p_A - \max_{L \in \mathcal{V}} p_L\right)
  \label{eq:vimplication}
\end{equation}
 
For mutex constraints (Type C), the violation $v_\perp$ is the
product of the probabilities of the two mutually exclusive events:
\begin{equation}
  v_\perp(A_1, A_2) = p_{A_1} \cdot p_{A_2}
  \label{eq:vmutex}
\end{equation}
 
The three rule types are combined into the FOL loss:
\begin{equation}
  \mathcal{L}_{\mathrm{FOL}} =
  \lambda_A\, \overline{v_\Rightarrow^{(A)}} +
  \lambda_B\, \overline{v_\Rightarrow^{(B)}} +
  \lambda_C\, \overline{v_\perp}
  \label{eq:lfol}
\end{equation}
The three rule families are weighted equally; the values of $\lambda_A$, $\lambda_B$ and $\lambda_C$ and the role of $\lambda_{\mathrm{FOL}}$ as the single tunable coefficient are stated in Section~\ref{sec:experimental_setup}.
 
The total training objective combines the supervised terms of the four heads, cross-entropy for the per-clip action head and the single-label high-level intention head and binary cross-entropy for the multi-label activity and low-level intention heads, restricted to the evaluation taxonomy, with the FOL regulariser:
\begin{equation}
  \mathcal{L}_{\mathrm{total}} =
  \mathcal{L}_{\mathrm{CE}} +
  \lambda_{\mathrm{FOL}}\,\mathcal{L}_{\mathrm{FOL}}
  \label{eq:ltotal}
\end{equation}
 
where $\mathcal{L}_{\mathrm{CE}}$ denotes the sum of the supervised
classification losses for the four output heads, and $\lambda_{\mathrm{FOL}}$
weights the contribution of the logic term relative to the
classification objective.
The empirical effect of $\lambda_{\mathrm{FOL}}$ is analysed
in Section~\ref{sec:fol_analysis}.
 
\begin{algorithm}[H]
\caption{NSGT Hierarchical Inference and Training}
\label{alg:nsgt}
\begin{algorithmic}[1]
\Require Fused multimodal features
  $f = [f_s;\, f_r;\, f_{ir};\, f_d]$;
  learnable biases $\beta_{\mathrm{trans}},\,
  \beta_{\mathrm{act}},\, \beta_{\mathrm{lli}}$;
  logic weight $\lambda_{\mathrm{FOL}}$.
\Ensure Predicted labels $\hat{a}, \hat{c}, \hat{l}, \hat{h}$;
  total loss $\mathcal{L}_{\mathrm{total}}$.
 
\Statex \textbf{Step 1: Input projection}
\State $z^{(0)} \gets
  \mathrm{LayerNorm}(\mathrm{Linear}(f)) +
  \mathrm{PosEncoding}$
 
\Statex \textbf{Step 2: HR-G bias (computed from first-pass predicted actions)}
\State $B_{ij} \gets
  \beta_{\mathrm{trans}}\,\mathbb{1}[\hat a_j \in
    \mathcal{N}_{\mathrm{trans}}(\hat a_i)] +
  \beta_{\mathrm{act}}\,\mathbb{1}[\mathcal{A}(\hat a_i) \cap
    \mathcal{A}(\hat a_j) \neq \emptyset] +
  \beta_{\mathrm{lli}}\,\mathbb{1}[\mathcal{L}(\hat a_i) \cap
    \mathcal{L}(\hat a_j) \neq \emptyset]$
\Statex \textbf{Step 2a (first pass):} $z^{(L)}_{1} \gets \mathrm{Encoder}(z^{(0)};\, B = 0)$;\; $\hat a^{(1)}_i \gets \arg\max_{a \in \mathcal{A}_{\mathrm{ret}}} \mathrm{Linear}_{\mathrm{act}}(z^{(L)}_{1})_{i,a}$, detached. Every $\hat a$ in $B_{ij}$ denotes $\hat a^{(1)}$; ground-truth labels are not used.
\State $B_{ii} \gets 0$ \Comment{diagonal zeroed after the weighted sum}
\Statex \textbf{Step 2b (second pass):} $B$ is added to the attention logits of every layer and head in Step~3.
\State $\beta_{\mathrm{trans}}, \beta_{\mathrm{act}}, \beta_{\mathrm{lli}} \gets \mathrm{clamp}(\beta_{\cdot},\, -3,\, 3)$ before forming $B$
 
\Statex \textbf{Step 3: Graph-structured reasoning}
\For{$\ell = 1$ \textbf{to} $L$}
  \State $z^{(\ell)} \gets
    \mathrm{Softmax}\!\left(
      \tfrac{QK^\top}{\sqrt{d_{\mathrm{model}}}} + B
    \right)V + \mathrm{FFN}(z^{(\ell-1)})$
  \Comment{same $B$ for all layers}
\EndFor
 
\Statex \textbf{Step 4: Multi-task hierarchical prediction}
\State $\hat{a} \gets \mathrm{Linear}_{\mathrm{act}}(z^{(L)})$
  \Comment{per-step action logits}
\State $z_{\mathrm{seq}} \gets
  \mathrm{MaskedMeanPool}(z^{(L)})$
\State $\hat{c} \gets
  \mathrm{Linear}_c(z_{\mathrm{seq}})$
  \Comment{activity logits}
\State $z_{\mathrm{lli}} \gets
  \mathrm{MLP}_{\mathrm{act \to lli}}(z_{\mathrm{seq}})$
\State $\hat{l} \gets
  \mathrm{Linear}_{\mathrm{lli}}(z_{\mathrm{lli}})$
  \Comment{LLI logits}
\State $\hat{h} \gets
  \mathrm{Linear}_{\mathrm{hli}}(
  \mathrm{MLP}_{\mathrm{lli \to hli}}(z_{\mathrm{lli}}))$
  \Comment{HLI logits}
 
\Statex \textbf{Step 5: Loss computation}
\State $\mathcal{L}_{\mathrm{CE}} \gets \mathrm{CE}(\hat{a}) + \mathrm{BCE}(\hat{c}) + \mathrm{BCE}(\hat{l}) + \mathrm{CE}(\hat{h})$
\Comment{BCE for the multi-label activity and LLI heads}
\State $\mathcal{L}_{\mathrm{FOL}} \gets
  \mathrm{SoftLogicViolation}(\hat{c}, \hat{l}, \hat{h})$
\State $\mathcal{L}_{\mathrm{total}} \gets
  \mathcal{L}_{\mathrm{CE}} +
  \lambda_{\mathrm{FOL}}\,\mathcal{L}_{\mathrm{FOL}}$
\State \Return
  $\{\hat{a}, \hat{c}, \hat{l}, \hat{h}\},\;
  \mathcal{L}_{\mathrm{total}}$
\end{algorithmic}
\end{algorithm}
 
\paragraph{Head dependencies and error propagation.}
The heads share the Transformer representation and are chained through continuous representations rather than through decisions. The pooled episode representation feeds the activity head directly and also feeds an MLP that produces the LLI hidden representation; that representation feeds the LLI head and, through a further MLP, the HLI head (equations~\eqref{eq:pooling} to \eqref{eq:hli_head}). The projection into the LLI representation is named for the levels it spans but takes the pooled representation as its input rather than the activity logits, and no discrete activity or LLI decision is passed to a later head. The
only place where a discrete prediction re-enters the computation is
the HR-G bias, which is indexed by first-pass predicted action identities, a property verified in Section~\ref{sec:hrg_logic}. An action error can therefore activate the wrong edges for
that clip, although the bias remains additive to the content attention
and bounded by clamping. The logic objective acts on the joint output
configuration during training, penalising higher-level predictions
that the lower levels cannot support. Two
empirical consequences are visible in the results. First, the
high-level intention is recovered far more often than the complete
lower-level label set: on the standard test split the proposed model
attains 78.41\% high-level macro-F1 against a joint exact-match
accuracy of 36.90\%, so a large fraction of episodes reach the correct
high-level intention from an imperfect activity or low-level-intention
set. Second, adding the lower-level heads to intention-only
supervision raises high-level macro-F1 from 67.89\% to 77.84\%
(Table~\ref{tab:head_ablation}), so lower-level evidence improves
rather than dominates the highest level.

%% ═══════════════════════════════════════════════════════════════════════════════
\section{The NTU-Hierarchy Dataset and Human Reasoning Graph}
\label{sec:dataset}
 
As surveyed in \cite{zhao2025deep}, existing intention recognition
benchmarks span text-only datasets (ATIS, CLINC150),
audio-visual datasets (MINE, MC-EIU), and domain-specific
multimodal resources.
None provides the combination of four-modality sensing and
four-level hierarchical annotation required for general-purpose
hierarchical intention reasoning.
To address this gap, NTU RGB+D~120 \cite{liu2019ntu} is used as
the base corpus for constructing a novel hierarchical benchmark.
By mapping its atomic action labels onto a structured four-level
ontological hierarchy and constructing semantically coherent
multi-clip episodes, a large-scale multimodal benchmark for
hierarchical intention reasoning is produced.
 
The manuscript describes the benchmark to the extent required for the dataset structure and the model evaluation to be self-contained. The design criteria of the benchmark, the mechanism by which each is realised, and the synthesis procedure are documented in a separate benchmark paper \cite{soleimani2026compositional} and are not restated here, so that the contribution of the present work, the NSGT and hierarchical intention recognition, remains in focus and the manuscript stays within length.
 
%%─────────────────────────────────────────────────────────────────
\subsection{Hierarchy Construction and Ontology Design}
\label{sec:hierarchy_construction}
%%─────────────────────────────────────────────────────────────────
 
A set of 58 action classes is selected from NTU RGB+D~120
\cite{liu2019ntu}, covering a broad range of daily-living and
social behaviours.
These base classes are organized into a four-level semantic
taxonomy, summarized in Table~\ref{tab:taxonomy}.
 
\begin{table}[t]
\centering
\caption{
  The four-level semantic taxonomy constructed from NTU RGB+D~120.
}
\label{tab:taxonomy}
\small
\begin{tabularx}{\columnwidth}{@{} l r X X @{}}
\toprule
\textbf{Level} & \textbf{Count} &
\textbf{Definition} & \textbf{Examples} \\
\midrule
Atomic Action    & 58 &
  Base kinematic labels for isolated motions. &
  \texttt{drink\_water}, \texttt{brushing\_teeth} \\
\addlinespace
Composite Activity & 39 &
  Coherent behavioural routines grouping related actions. &
  \texttt{PersonalCare}, \texttt{DrinkLiquid} \\
\addlinespace
Low-Level Intention (LLI) & 28 &
  Intermediate short-term goals within an episode. &
  \texttt{Eating}, \texttt{DocumentTasks} \\
\addlinespace
High-Level Intention (HLI) & 8 &
  Broad, long-term cognitive objectives. &
  \texttt{MorningRoutine}, \texttt{OfficeWorkSession} \\
\bottomrule
\end{tabularx}
\end{table}
 
\textbf{Episode construction.}
Activity-level episodes are synthesized via a probabilistic
transition graph (described in Section~\ref{sec:hrg_logic}).
Each episode comprises 4 to 17 action clips.
The total number of activity-level episodes, which constitute
the model's training samples, is 11,436 across subject-disjoint
train, validation, and test splits, with a further 3,415
episodes held out in a compositional test set, as reported in
Table~\ref{tab:dataset}.
 
\textbf{Subject consistency constraint.}
A strict subject consistency constraint is enforced during
episode construction: all clips within a single episode are
drawn exclusively from the same subject identity.
This ensures that the inferred intention reflects a
single performer within each assembled episode rather than an artifact
of combining different identities. It does not make the assembled clips
a naturally recorded continuous trajectory.
 
\textbf{Multimodal features.}
Each clip is represented across all four sensor modalities
(skeleton, RGB, infrared, depth).
Pre-extracted features are concatenated into a unified
5120-dimensional vector
(2048 skeleton + 1024 RGB + 1024 IR + 1024 depth).
Encoder architectures and extraction protocols are detailed
in Section~\ref{sec:features}.
 
\textbf{Synthesized labels.}
The hierarchical labels at the activity, LLI, and HLI levels
and the FOL constraints are both derived from the designed ontology.
Label assignment and loss computation occur at separate stages, but
they are not conceptually independent sources of semantic knowledge,
as discussed in Section~\ref{sec:hrg_logic}.
 
\begin{table}[t]
\centering
\caption{
  Dataset statistics for the NTU-Hierarchy benchmark.
  Activity-level episodes are the training samples. Splits are
  subject-disjoint by construction; the compositional test set holds out
  unseen intention-level combinations and, by design, shares performers
  with training.
}
\label{tab:dataset}
\small
\begin{tabular}{lrrr}
\toprule
\textbf{Split} &
\textbf{Episodes} &
\textbf{Avg.\ length} &
\textbf{Performers} \\
 & & \textbf{(clips)} & \\
\midrule
Train                 & 7,062  & 6.69 & 42 \\
Validation            & 1,519  & 6.90 & 6  \\
Test                  & 2,855  & 7.04 & 12 \\
\midrule
Total (subject-disjoint) & 11,436 & 6.80 & 60 \\
\midrule
Compositional test    & 3,415  & 7.36 & 42$^{\dagger}$ \\
\bottomrule
\end{tabular}
 
\medskip
 
\begin{tabular}{lr}
\toprule
\textbf{Property} & \textbf{Value} \\
\midrule
Hierarchy classes (Action / Activity / LLI / HLI) & 58 / 39 / 28 / 8 \\
Sensor modalities                                  & 4 \\
Episode length range (clips)                       & 4--17 \\
Split protocol                                     & subject-disjoint \\
FOL training constraints (Type~A / Type~B / Type~C) & 39 / 28 / 420 \\
\bottomrule
\end{tabular}
 
\smallskip
{\footnotesize $^{\dagger}$The compositional test set shares performers with
training by design; its purpose is to test generalization to unseen
intention-level combinations rather than to unseen performers, so its
performers are not counted in the subject-disjoint total.}
\end{table}
 
%%─────────────────────────────────────────────────────────────────
\subsection{Multi-Membership and Disjunctive Rule Design}
\label{sec:ambiguity}
%%─────────────────────────────────────────────────────────────────
 
A deliberate design feature of the hierarchy is the preservation
of \textbf{one-to-many ontological relationships}.
The LLI \texttt{DrinkBreak}, for example, is a valid
member of four of the eight HLIs, because a short drink break
serves multiple long-term goals depending on temporal context.
Similarly, the activity \texttt{WalkAway} maps validly to both
\texttt{FriendlyDeparture} and \texttt{HostileAftermath},
depending on the broader episode context.
This is not a labeling ambiguity but a reflection of genuine
behavioral flexibility: the same activity sequence can serve
different high-level purposes depending on surrounding context
(Figure~\ref{fig:hrg_hierarchy}).
 
This one-to-many structure directly motivates the disjunctive
rule formulation in the FOL loss.
A deterministic mapping of the form
$\mathrm{Activity}(X, A) \Rightarrow \mathrm{LLI}(X, L)$
would be incorrect for activities with multiple valid LLI
pathways.
Instead, the disjunctive form is adopted:
\begin{equation}
  \mathrm{Activity}(X, A)
  \;\Rightarrow\;
  \bigvee_{L \in \mathcal{V}(A)} \mathrm{LLI}(X, L)
  \label{eq:disjunctive}
\end{equation}
where $\mathcal{V}(A)$ is the set of all valid LLIs for
activity $A$, encoded in the binary mapping matrix
$\mathbf{R}_{\mathrm{act \to lli}} \in \{0,1\}^{39 \times 28}$.
The loss evaluates this rule by checking whether the maximum
predicted probability across all valid consequents,
$\max_{L \in \mathcal{V}(A)} p_L$, is sufficiently high,
rewarding the model for predicting any valid LLI pathway rather
than a single fixed target.
This formulation is fully differentiable and connects the
ontology directly to the loss described in
Section~\ref{sec:fol_loss}.
 
This one-to-many design makes a direct action-to-HLI lookup
non-unique and motivates the deterministic lookup baseline evaluated
in Section~\ref{sec:lookup}. The reported lookup implementation is
substantially weaker than the learned model at the intention levels;
its exact resolution and tie-breaking rules are documented there.
 
%%─────────────────────────────────────────────────────────────────
\subsection{Human Reasoning Graph and FOL Rule Construction}
\label{sec:hrg_logic}
%%─────────────────────────────────────────────────────────────────
 
The HR-G serves two distinct roles in the framework.
Its transition graph component is used during dataset
construction to produce temporally coherent episode sequences.
Its FOL rule component is used during training as a constraint
on the model's predictions.
These two roles are implemented at different stages, but both are
derived from the same ontology.
 
\textbf{Relation between labels and logic constraints.}
A circular supervision problem arises when the labels used
for training are derived from the same rules that the model
is being trained to satisfy.
In such a setup, the model can achieve a high hierarchy consistency
score simply by reproducing the labeling logic, rather than
learning genuine hierarchical reasoning from the data.
 
In the proposed framework, episode labels are assigned before model
training, and the differentiable FOL term is applied only to model
outputs. This operational separation prevents the loss implementation
from directly reading target labels. It does not, however, make labels
and constraints independent, because both encode relations from the
same ontology. An increase in hierarchy consistency after adding the FOL term is therefore not interpreted as independent evidence of reasoning.
Evidence beyond this objective-aligned diagnostic is sought
in per-level predictive metrics, the no-FOL graph comparison, the
lookup baseline, and the compositional split.

\textbf{Scope of the hierarchy-consistency metric.}
The logic loss operates during training on continuous probability
vectors, whereas the reported hierarchy-consistency rate is a discrete
post-hoc check on predicted parent--child relations. The two are not
identical computations, but they are aligned because they use the same
ontology relations. The rate is therefore reported as a diagnostic of
cross-level coherence, not as an independent validation of the FOL
objective. Applying the same metric to models trained without the logic
term remains useful for controlled comparison, but does not remove this
conceptual dependence.

\textbf{Evaluation-only semantic audit.} Separately from the hierarchy consistency rate and from the 487 active constraints used by the FOL training loss, a set of 25 first-order semantic rules over the ontology serves as an evaluation instrument. Group X contains the three rules that are also enforced during episode synthesis. Group Y contains the remaining 22 rules, which are accessed neither during episode generation nor during training. Label construction and FOL training are operationally separated, but they are not conceptually independent, because both rely on the same ontology. What the separation supports is therefore narrower, and is stated as such: the training loss implementation never references the rule table, a property enforced by an automated check, and on the ground-truth labels of the test split the Group Y rules are violated at 9.3\% macro-averaged over rules and 5.0\% instance-weighted, against zero for Group X, which is empirical evidence that the Group Y rules were not enforced by construction. Per-rule violation rates of the evaluated models, for the 10 Group Y rules computable from episode-level predictions, are reported in the supplementary material, together with the 12 rules that require segment-level or order information, or that describe completeness or ordering not represented at episode level, and are therefore not computable in this setting. Across the compared conditions the trained models violate the Group Y rules at 10.8 to 11.1\% macro-averaged and 4.8 to 4.9\% instance-weighted on the test split, close to the ground-truth rates. The logic objective leaves the Group Y rules essentially unchanged, which is consistent with its scope being the 487 active constraints rather than a general semantic oracle.

\textbf{Probabilistic transition graph.}
A directed, weighted transition graph is defined over the action classes of the ontology,
encoding which actions naturally follow which
in a temporally ordered sequence.
The edge weights are asymmetric, reflecting realistic behavioral ordering:
the transition probability from an approach action to a greeting
is high, while the reverse ordering is near zero.
The learned HR-G bias scalars (reported in
Section~\ref{sec:graph_impact}) quantify the relative contribution of
the three predefined edge types under the tested setting, with the
largest coefficient assigned to low-level-intention co-membership.
 
The graph produces temporally ordered episodes whose labels and
training-time constraints are both derived from the ontology. Their
construction procedures are separate, but their semantic source is
shared.
 
\textbf{FOL rules as training-time constraints.}
A total of 487 active constraints are applied as training-time penalties: 39
disjunctive Type-A activity$\to$LLI implications, 28 disjunctive
Type-B LLI$\to$HLI implications, and 420 Type-C mutual exclusion
constraints.
These rules penalize ontological inconsistencies in the model's output
predictions. Because labels and constraints share the ontology, their
effects are interpreted through controlled ablations rather than as
independent supervision sources.
 
Mutual exclusion constraints are applied conservatively: a
mutex penalty is imposed only between activity pairs whose
sets of valid LLIs are entirely disjoint.
This conservative design prevents false conflicts between activities
that share valid intention pathways at higher levels of the hierarchy,
and is the reason the synthesized label sets
satisfy every Type-C constraint by construction: each episode carries
a single high-level intention, so all of its activities share at
least that intention and no mutually exclusive pair can co-occur.
 
\textbf{Division of temporal and ontological reasoning.}
The transition graph enforces directed temporal ordering
during episode construction through its asymmetric edge
weights.
The FOL rules, by contrast, operate on episode-level
predictions: after mean-pooling produces a single probability
vector over the episode, the logic constraints verify
ontological consistency regardless of internal clip ordering.
 
Temporal ordering is captured by the Transformer's positional
encodings and the asymmetric HR-G transition bias
$\beta_{\mathrm{trans}}$.
Ontological consistency is promoted by the FOL loss.
These two mechanisms address complementary aspects of
hierarchical reasoning: the model is penalised for
ontological contradictions by the FOL loss, and learns
temporal ordering through the attention mechanism.
 
\textbf{Learned scalar parameters.}
The three learned scalars $\beta_{\mathrm{trans}}$,
$\beta_{\mathrm{act}}$, and $\beta_{\mathrm{lli}}$ modulate
the HR-G attention bias in every Transformer layer, allowing
the model to discover which graph edge types are most
informative.
Their converged values and the implications for model
interpretability are analyzed in
Section~\ref{sec:graph_impact}.

%% ── TikZ figure placeholder ────────────────────────────────────────────────
 
\textbf{Label independence of the attention bias at inference.}
The bias $B$ is constructed by a two-pass procedure that uses no label information at inference. In the first pass the reasoning network is run over the precomputed clip features without any attention bias, and the action head yields a prediction for every clip slot, restricted to the 58 retained action classes and detached from the computational graph. In the second pass these predicted action identities index the successor and co-membership tables to build $B$, which is then added to the attention logits of every layer and head; the diagonal of $B$ is set to zero after the weighted sum and padded positions contribute nothing. The forward function of the model receives only the clip features and the padding mask, so ground-truth action, activity or intention labels cannot reach the bias construction. This property is verified by a corruption-invariance test in which every ground-truth field of an evaluation batch is replaced by random labels: the first-pass action logits, the second-pass action logits and the activity, low-level intention and high-level intention logits are identical, element for element, between the clean and the corrupted batches.
 
\begin{figure*}[t]
  \centering
  \resizebox{\textwidth}{!}{%
  \begin{tikzpicture}[
      font=\small,
      >=stealth,
      %% ── node styles per layer ─────────────────────────────
      act/.style={
          draw=black!40, fill=black!5, rounded corners=2pt,
          minimum width=22mm, minimum height=6mm, align=center,
          font=\scriptsize\ttfamily},
      atv/.style={
          draw=blue!55, fill=blue!8, rounded corners=2pt,
          minimum width=26mm, minimum height=7mm, align=center,
          font=\scriptsize},
      lli/.style={
          draw=teal!70, fill=teal!10, rounded corners=2pt,
          minimum width=26mm, minimum height=7mm, align=center,
          font=\scriptsize},
      hli/.style={
          draw=orange!80, fill=orange!12, rounded corners=2pt,
          minimum width=30mm, minimum height=8mm, align=center,
          font=\scriptsize\bfseries},
      %% ── edge styles ───────────────────────────────────────
      arr/.style={draw=black!55, -{Stealth[length=1.6mm]}, line width=0.35pt},
      arr_share/.style={draw=purple!70, -{Stealth[length=1.8mm]}, line width=0.7pt},
      arr_fan/.style={draw=orange!85, -{Stealth[length=1.8mm]}, line width=0.7pt},
      arr_example/.style={draw=orange!85, -{Stealth[length=2mm]},
                          line width=1.2pt, dashed},
      %% ── column x-coordinates ─────────────────────────────
      xA/.initial=0,        %% Layer 1 (actions)
      xG/.initial=4.2,      %% Layer 2 (activities)
      xL/.initial=8.6,      %% Layer 3 (LLIs)
      xH/.initial=13.0      %% Layer 4 (HLIs)
    ]

    %% ── Column headers ───────────────────────────────────────
    \node[font=\small\bfseries, anchor=south]
      at (0, 0.6) {Layer 1: Action};
    \node[font=\small\bfseries, anchor=south, blue!60!black]
      at (4.2, 0.6) {Layer 2: Activity};
    \node[font=\small\bfseries, anchor=south, teal!60!black]
      at (8.6, 0.6) {Layer 3: LLI};
    \node[font=\small\bfseries, anchor=south, orange!60!black]
      at (13.0, 0.6) {Layer 4: HLI};
 
    %% ── Layer 1: Actions (grey), stacked vertically ─────────
    \node[act] (a01) at (0, 0.0)   {drink water};
    \node[act] (a02) at (0,-0.7)   {open bottle};
    \node[act] (a03) at (0,-1.4)   {cheers and drink};
    \node[act] (a04) at (0,-2.1)   {eat meal};
    \node[act] (a05) at (0,-2.8)   {brushing teeth};
    \node[act] (a06) at (0,-3.5)   {brushing hair};
    \node[act] (a07) at (0,-4.2)   {wipe face};
    \node[act] (a08) at (0,-4.9)   {reading};
    \node[act] (a09) at (0,-5.6)   {writing};
    \node[act] (a10) at (0,-6.3)   {running on spot};
    \node[act] (a11) at (0,-7.0)   {sneeze/cough};
    \node[act] (a12) at (0,-7.7)   {handshaking};
    \node[act] (a13) at (0,-8.4)   {point finger at other};
    \node[act] (a14) at (0,-9.1)   {phone call};
 
    %% ── Layer 2: Activities (blue) ───────────────────────────
    \node[atv] (g_drink)   at (4.2,-0.7)  {DrinkLiquid};
    \node[atv] (g_eat)     at (4.2,-2.1)  {EatFood};
    \node[atv] (g_care)    at (4.2,-3.5)  {PersonalCare};
    \node[atv] (g_rw)      at (4.2,-5.25) {ReadWrite};
    \node[atv] (g_run)     at (4.2,-6.3)  {RunInPlace};
    \node[atv] (g_resp)    at (4.2,-7.0)  {RespiratoryDistress};
    \node[atv] (g_phg)     at (4.2,-7.7)  {PhysicalGreet};
    \node[atv] (g_conf)    at (4.2,-8.4)  {Confront};
    \node[atv] (g_ph)      at (4.2,-9.1)  {PhoneCall};
 
    %% ── Layer 3: LLIs (teal) ─────────────────────────────────
    \node[lli] (l_lb)  at (8.6,-1.2)  {LightBreakfast};
    \node[lli] (l_eat) at (8.6,-2.4)  {Eating};
    \node[lli] (l_doc) at (8.6,-3.6)  {DocumentTasks};
    \node[lli] (l_ex)  at (8.6,-4.8)  {MainExercise};
    \node[lli] (l_sy)  at (8.6,-5.8)  {SymptomOnset};
    \node[lli] (l_gr)  at (8.6,-6.9)  {GreetingRitual};
    \node[lli] (l_cf)  at (8.6,-7.9)  {Confrontation};
    \node[lli] (l_st)  at (8.6,-8.8)  {SettleIn};
    \node[lli] (l_pc)  at (8.6,-9.7)  {PhoneCheck};
 
    %% ── Layer 4: HLIs (orange) ───────────────────────────────
    \node[hli] (h_mr)  at (13.0,-0.7)  {MorningRoutine};
    \node[hli] (h_mc)  at (13.0,-2.1)  {MealConsumption};
    \node[hli] (h_ow)  at (13.0,-3.5)  {OfficeWorkSession};
    \node[hli] (h_pe)  at (13.0,-4.9)  {PhysicalExercise};
    \node[hli] (h_me)  at (13.0,-6.0)  {MedicalEpisode};
    \node[hli] (h_fs)  at (13.0,-7.1)  {FriendlySocial};
    \node[hli] (h_hs)  at (13.0,-8.2)  {HostileSocial};
    \node[hli] (h_rh)  at (13.0,-9.3)  {ReturnHome};
 
    %% ── Layer 1 -> Layer 2 edges (parent tree) ───────────────
    \draw[arr] (a01.east) -- (g_drink.west);
    \draw[arr] (a02.east) -- (g_drink.west);
    \draw[arr] (a03.east) -- (g_drink.west);
    \draw[arr] (a05.east) -- (g_care.west);
    \draw[arr] (a06.east) -- (g_care.west);
    \draw[arr] (a07.east) -- (g_care.west);
    \draw[arr] (a08.east) -- (g_rw.west);
    \draw[arr] (a09.east) -- (g_rw.west);
    \draw[arr] (a10.east) -- (g_run.west);
    \draw[arr] (a11.east) -- (g_resp.west);
    \draw[arr] (a12.east) -- (g_phg.west);
    \draw[arr] (a13.east) -- (g_conf.west);
    \draw[arr] (a14.east) -- (g_ph.west);
 
    %% Example-path edge from action layer (dashed, orange) starts here
    \draw[arr_example] (a04.east) -- (g_eat.west);
 
    %% ── Layer 2 -> Layer 3 edges ─────────────────────────────
    \draw[arr] (g_drink.east) -- (l_lb.west);
    \draw[arr] (g_drink.east) -- (l_eat.west);
    \draw[arr] (g_care.east)  -- (l_eat.west);
    %% Sharing: EatFood feeds both LightBreakfast and Eating (highlighted)
    \draw[arr_share] (g_eat.east) -- (l_lb.west);
    \draw[arr_share] (g_eat.east) -- (l_eat.west);
    \draw[arr] (g_rw.east)    -- (l_doc.west);
    \draw[arr] (g_run.east)   -- (l_ex.west);
    \draw[arr] (g_resp.east)  -- (l_sy.west);
    \draw[arr] (g_phg.east)   -- (l_gr.west);
    \draw[arr] (g_conf.east)  -- (l_cf.west);
    \draw[arr] (g_ph.east)    -- (l_pc.west);
    \draw[arr] (g_ph.east)    -- (l_doc.west);
 
    %% ── Layer 3 -> Layer 4 edges ─────────────────────────────
    \draw[arr] (l_lb.east)  -- (h_mr.west);
    \draw[arr] (l_eat.east) -- (h_mc.west);
    \draw[arr] (l_doc.east) -- (h_ow.west);
    \draw[arr] (l_ex.east)  -- (h_pe.west);
    \draw[arr] (l_sy.east)  -- (h_me.west);
    \draw[arr] (l_gr.east)  -- (h_fs.west);
    \draw[arr] (l_cf.east)  -- (h_hs.west);
    \draw[arr] (l_st.east)  -- (h_rh.west);
    %% Fan-out: PhoneCheck reaches four HLIs (highlighted)
    \draw[arr_fan] (l_pc.east) -- (h_mr.west);
    \draw[arr_fan] (l_pc.east) -- (h_mc.west);
    \draw[arr_fan] (l_pc.east) -- (h_ow.west);
    \draw[arr_fan] (l_pc.east) -- (h_rh.west);
 
    %% Example-path continuation (dashed, orange) through the tree
    \draw[arr_example] (g_eat.east) .. controls +(1.0,0) and +(-1.0,0) .. (l_lb.west);
    \draw[arr_example] (l_lb.east)  -- (h_mr.west);
 
    %% ── Footer counts ────────────────────────────────────────
    \node[font=\scriptsize, anchor=north]
      at (0,  -10.0) {$|\mathcal{A}|=58$};
    \node[font=\scriptsize, anchor=north, blue!60!black]
      at (4.2,-10.0) {$|\mathcal{G}|=39$};
    \node[font=\scriptsize, anchor=north, teal!60!black]
      at (8.6,-10.0) {$|\mathcal{L}|=28$};
    \node[font=\scriptsize, anchor=north, orange!60!black]
      at (13.0,-10.0) {$|\mathcal{H}|=8$};
 
    %% ── Legend (bottom-centred) ──────────────────────────────
    \node[
        draw=black!30, fill=white, rounded corners=3pt,
        font=\scriptsize, align=left, inner sep=5pt
      ] at (6.6,-11.2) {%
      \tikz\draw[arr_example, line width=1.2pt] (0,0)--(0.7,0);~%
      Example path:
      \texttt{eat\ meal}
      $\to$ \texttt{EatFood}
      $\to$ \texttt{LightBreakfast}
      $\to$ \texttt{MorningRoutine}\\[2pt]
      \tikz\draw[arr_share, line width=0.7pt] (0,0)--(0.7,0);~%
      Shared activity: \texttt{EatFood} appears under two LLIs
      that resolve to different HLIs\\[2pt]
      \tikz\draw[arr_fan, line width=0.7pt] (0,0)--(0.7,0);~%
      Fan-out: \texttt{PhoneCheck} is reachable under four HLIs
    };
 
  \end{tikzpicture}%
  }
  \caption{%
    The four-level Human Reasoning Graph (HR-G) over the
    NTU-Hierarchy evaluation universe:
    58 atomic actions, 39 composite activities,
    28 Low-Level Intentions (LLI), and 8 High-Level Intentions (HLI).
    Layer~1 (grey): atomic actions.
    Layer~2 (blue): composite activities.
    Layer~3 (teal): LLIs.
    Layer~4 (orange): HLIs.
    A representative subgraph is shown; every drawn edge is taken
    from the ontological parent tree, and every drawn node belongs
    to the evaluation taxonomy of Table~\ref{tab:taxonomy}.
    The dashed orange path traces how \texttt{eat\ meal} is
    propagated through \texttt{EatFood} and \texttt{LightBreakfast}
    to \texttt{MorningRoutine}.
    Several mappings are deliberately non-injective:
    the activity \texttt{EatFood} is shared by two LLIs that resolve
    to different HLIs (\texttt{MorningRoutine} vs.
    \texttt{MealConsumption}),
    and the intention \texttt{PhoneCheck} is reachable under four
    distinct HLIs.
    The non-injective mappings motivate the lookup baseline evaluated
    in Section~\ref{sec:lookup} and require an explicit rule for
    resolving multiple valid parents.
  }
  \label{fig:hrg_hierarchy}
\end{figure*}
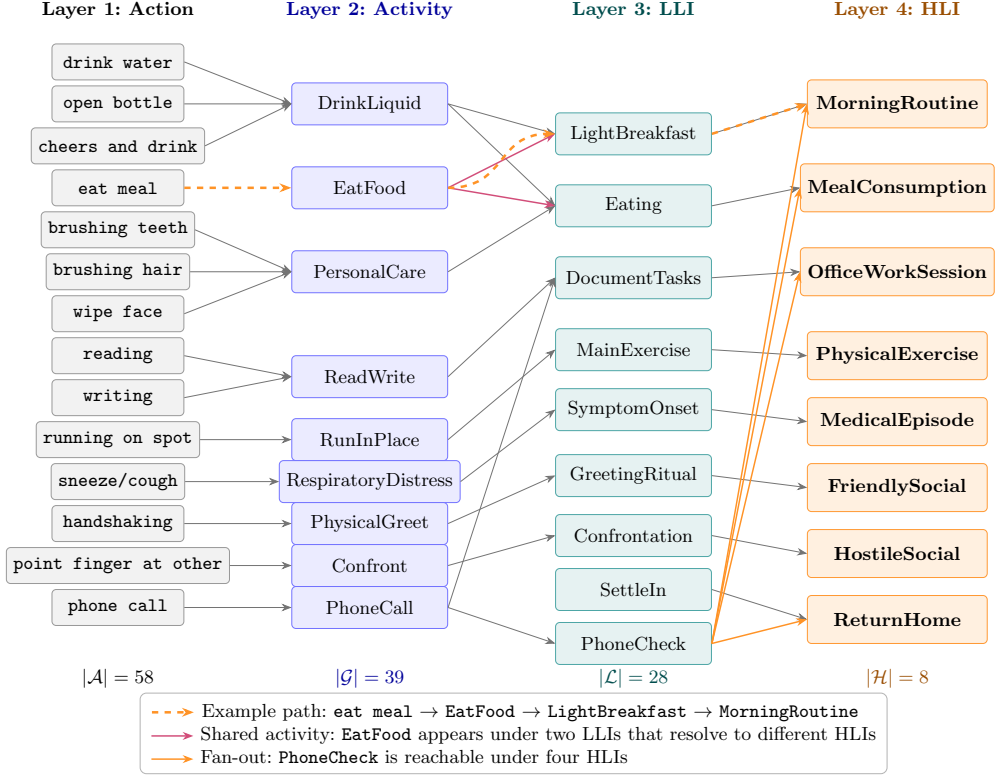

\textbf{Visualization.}
A representative slice of the HR-G ontology is illustrated in
Figure~\ref{fig:hrg_hierarchy}.
The four levels comprise 58 action, 39 activity, 28 LLI, and 8
HLI classes; the figure shows a curated subset of nodes and
edges to make the multi-parent structure legible, with directed
edges representing valid ontological transitions.
An ontology path such as \texttt{handshaking} can be traced
through \texttt{ACT\_PhysicalGreet} $\to$
\texttt{LLI\_GreetingRitual} $\to$
\texttt{HLI\_FriendlySocial}. This makes the designed hierarchy
inspectable, but the ontology diagram alone is not a model-specific
audit trail. A concrete prediction trace is discussed separately in
Section~\ref{sec:audit}.

%% ═══════════════════════════════════════════════════════════════════════════
\section{Experimental Setup}\label{sec:experimental_setup}
 
\textbf{Fairness of comparisons.}
All models are trained and evaluated under controlled conditions. Every
model receives the same
5120-dimensional multimodal features from the same frozen encoders, is
trained and tested on the same subject-disjoint splits, and follows the
same training protocol, with no per-model hyperparameter tuning and
model selection on the validation LLI macro-F1 for every condition.
This controls data, features, splits, and training budget. Because no
equal per-model hyperparameter search was performed, the baselines are
presented as controlled implementations rather than fully optimized
representatives of their model families.
 
\textbf{Feasibility under constrained hardware.}
Because the perceptual encoders are precomputed and frozen, the reported
resource measurements concern only the reasoning module, which has
approximately 3.6 million trainable parameters. Table~\ref{tab:efficiency}
reports inference latency, throughput, forward-pass allocated memory, and
lower-bound operation counts on a Tesla T4. These measurements support
feasibility on that setup but do not establish a minimum hardware
requirement or peak end-to-end training memory. More constrained devices
may require a smaller batch size or reduced precision.
 
\begin{table}[t]
\centering
\caption{
  Efficiency of the reasoning models on a single NVIDIA Tesla T4, measured at seed 42
  from the trained checkpoints. Parameters are total counts. FLOPs are reported as
  multiply-add operations counted by fvcore for a padded episode of length 32; this count
  covers the linear and projection operations and excludes fused attention, the recurrent
  cell, and scatter-based graph aggregation, so the values for the attention, recurrent,
  and graph models are lower bounds. The proposed model is measured over its full two-pass
  inference. Latency is the median and 95th percentile over 200 timed iterations at batch
  size 1; throughput is measured at batch size 32; peak memory is the maximum allocated
  during a batch-size-32 forward. Latency and memory are specific to the T4.
}
\label{tab:efficiency}
\small
\setlength{\tabcolsep}{4pt}
\begin{tabular}{lrrrrrr}
\toprule
\textbf{Model} & \textbf{Params} & \textbf{FLOPs} &
\textbf{Lat. p50} & \textbf{Lat. p95} & \textbf{Thrpt} & \textbf{Mem} \\
 & (M) & (MMACs) & (ms) & (ms) & (ep/s) & (MB) \\
\midrule
NSGT+FOL (two-pass) & 3.60 & 178.4 & 14.5 & 35.4 & 4409 & 133.8 \\
HT (no FOL)         & 3.60 & 110.3 & 3.0 & 3.7 & 10450 & 130.9 \\
HT+FOL              & 3.60 & 110.3 & 2.9 & 3.4 & 9942 & 130.9 \\
Seq-Transformer     & 3.07 & 93.3 & 1.6 & 1.8 & 14317 & 130.9 \\
MLP                 & 2.93 & 89.0 & 0.9 & 1.0 & 18383 & 124.7 \\
LSTM                & 2.35 & 45.0 & 2.8 & 3.4 & 9620 & 167.9 \\
R-GCN               & 4.20 & $23.2 \pm 6.7$ & 7.4 & 9.0 & 3395 & 73.3 \\
\bottomrule
\end{tabular}
\end{table}
 
\label{sec:setup}
 
%%─────────────────────────────────────────────────────────────────
\subsection{Implementation Details}
\label{sec:impl}
%%─────────────────────────────────────────────────────────────────
 
All models are implemented in PyTorch and trained on a single
NVIDIA Tesla T4 GPU (16\,GB VRAM).
The NSGT reasoning modules are trained from scratch; all feature
encoders remain frozen throughout training.
 
\textbf{Architecture configuration.}
The NSGT and HT backbones share the same hyperparameters:
model dimension $d_{\mathrm{model}} = 256$,
number of attention heads $n_{\mathrm{heads}} = 8$,
number of Transformer layers $L = 4$,
feed-forward dimension $d_{\mathrm{ff}} = 512$,
and dropout rate 0.1.
The total number of trainable parameters in the full NSGT model is
3,598,261. Removing the three $\beta$ scalars yields the
uniform-attention HT backbone with 3,598,258 parameters; the FOL
objective adds no parameters, so all conditions of a given backbone
are parameter-identical.
The HR-G bias scalars are initialised to
$\beta_{\mathrm{trans}} = 0.1$,
$\beta_{\mathrm{act}} = 0.1$, and
$\beta_{\mathrm{lli}} = 0.05$.
 
\textbf{Optimiser.}
AdamW \cite{loshchilov2017decoupled} is used with an initial
learning rate $\mathrm{lr} = 2 \times 10^{-4}$,
weight decay $10^{-4}$, and cosine annealing to a minimum of
$10^{-6}$ over $T_{\max} = 50$ epochs.
Gradients are clipped to a maximum $\ell_2$ norm of 1.0.
These optimisation components are standard, robust defaults rather than dataset-specific choices: cosine annealing gives smooth convergence without a hand-designed step schedule, clipping stabilises the larger early gradients introduced by the logic term, and the identical schedule is applied to every model and baseline without per-model re-tuning.

\textbf{Loss function.}
The total training loss combines a hierarchical cross-entropy
term and the differentiable FOL regularisation term:
\begin{equation}
  \mathcal{L}_{\mathrm{total}}
  = \mathcal{L}_{\mathrm{CE}}
  + \lambda_{\mathrm{FOL}}\,\mathcal{L}_{\mathrm{FOL}}
  \label{eq:total_loss}
\end{equation}
where $\mathcal{L}_{\mathrm{CE}}$ is the sum of cross-entropy
losses across all four prediction heads, and
$\mathcal{L}_{\mathrm{FOL}}$ acts as a soft regulariser,
penalising ontological contradictions without overriding the
data-driven learning signal.
Within $\mathcal{L}_{\mathrm{FOL}}$, the three rule families are weighted equally
($\lambda_A = \lambda_B = \lambda_C = 1$), so the
logic weight $\lambda_{\mathrm{FOL}}$ is the single tunable coefficient
of the logic term.
Two values of the logic weight are evaluated:
$\lambda_{\mathrm{FOL}} = 0.1$, which yields a small consistency
gain with negligible change in classification accuracy, and
$\lambda_{\mathrm{FOL}} = 0.5$, which maximises logic
consistency while action F1 remains essentially unchanged.
The empirical effect of the logic weight is analyzed in
Section~\ref{sec:fol_analysis}.
 
\textbf{Early stopping.}
Training is halted if validation LLI macro-F1 does not improve
for 10 consecutive epochs.
The checkpoint with the highest validation LLI macro-F1 is
retained for test evaluation.
LLI macro-F1 is selected as the stopping criterion because it
represents the primary intention-recognition objective and
occupies the middle of the four-level hierarchy, providing a
signal that reflects both low-level recognition quality and
high-level reasoning ability.
 
\textbf{Batch size and sequence handling.}
A batch size of 32 is used throughout.
Episodes vary in length; variable-length sequences
are zero-padded to the maximum length within each batch.
A boolean padding mask is passed to all attention layers so
that padded positions are excluded from both attention
computation and mean-pooling.
 
\textbf{Reproducibility and statistical protocol.}
The principal model comparison is trained independently with five random
seeds (42 to 46), each seed fixing all stochastic components (NumPy,
PyTorch, and Python \texttt{random}). The head and modality ablations
use three seeds and are identified as secondary studies. Main tables
report means and standard deviations across the corresponding runs. The
data splits are fixed, subject-disjoint, and identical across all
conditions (Section~\ref{sec:hierarchy_construction}), so that every
model is trained and evaluated on exactly the same partition.
 
%%─────────────────────────────────────────────────────────────────
\subsection{Feature Extraction}
 
\textbf{Rationale for frozen encoders.}
The perceptual encoders are used as fixed feature extractors to provide
the same feature substrate to every reasoning model. This controls
feature learning within the reported comparison. The modality ablation
also shows that no evaluated single stream trivially solves the task
(the strongest evaluated single stream reaches 79.60\% action top-1,
whereas skeleton reaches 55.41\%). Fine-tuning could change both
absolute performance and the interaction between perception and
reasoning; it was not evaluated and is left as future work.
 
\label{sec:features}
%%─────────────────────────────────────────────────────────────────
 
All multimodal features are extracted offline using frozen
pretrained encoders and stored as consolidated tensors,
removing encoder computation from the training loop.
This design ensures that reported training times and
cross-model comparisons reflect only the reasoning module.
 
\begin{itemize}
  \item \textbf{Skeleton (2048-dim):}
    PoseC3D \cite{duan2022revisiting} with a SlowOnly-R50
    backbone, pretrained on NTU RGB+D~120 under the
    cross-subject protocol.
    Each clip is represented as a 3D heatmap volume, yielding
    a 2048-dimensional feature vector.
 
  \item \textbf{RGB (1024-dim):}
    VideoMAE-v2 \cite{wang2023videomae} with a ViT-Large
    backbone, pretrained on Kinetics-700.
    Sixteen-frame windows at the native frame rate are encoded
    into 1024-dimensional clip features.
 
  \item \textbf{Infrared (1024-dim):}
    VideoMAE-v2 fine-tuned on infrared frames from NTU
    RGB+D~120, producing 1024-dimensional features.
    This modality captures thermal body shape under low-light
    conditions where RGB appearance degrades.
 
  \item \textbf{Depth (1024-dim):}
    DINOv2 \cite{oquab2023dinov2} applied to depth-map frames,
    producing 1024-dimensional features.
    Depth provides 3D structural information complementary to
    the 2D appearance captured by RGB.
\end{itemize}
 
The four modality vectors are concatenated per clip to form a
5120-dimensional fused feature:
\begin{equation}
  \mathbf{f}
  = \bigl[\mathbf{f}_s;\,\mathbf{f}_r;\,\mathbf{f}_{ir};\,
    \mathbf{f}_d\bigr]
  \in \mathbb{R}^{5120}
  \label{eq:fused}
\end{equation}
No explicit dimensionality reduction is applied before the
linear projection layer.
Table~\ref{tab:features} summarises the encoder configurations.
 
\begin{table}[t]
\centering
\caption{
  Multimodal feature encoders.
  All encoders are frozen during training.
}
\label{tab:features}
\small
\begin{tabular}{llrl}
\toprule
\textbf{Modality} & \textbf{Encoder} &
\textbf{Dim} & \textbf{Pretrain data} \\
\midrule
Skeleton & PoseC3D SlowOnly-R50  & 2048 & NTU RGB+D 120 \\
RGB      & VideoMAE-v2 ViT-L     & 1024 & Kinetics-700  \\
Infrared & VideoMAE-v2 (IR FT)   & 1024 & NTU RGB+D 120 \\
Depth    & DINOv2 ViT-B/14       & 1024 & LVD-142M      \\
\bottomrule
\end{tabular}
\end{table}
 
%%─────────────────────────────────────────────────────────────────
\subsection{Baselines}
\label{sec:baselines}
%%─────────────────────────────────────────────────────────────────
 
The NSGT is compared against four baselines, a multilayer perceptron,
a bidirectional LSTM, a sequence Transformer, and an external
relational graph convolutional network, together with one ablated
variant of the proposed architecture, all trained with the same fused
5120-dimensional features, dataset splits, optimiser, and evaluation
protocol.
The four baselines do not incorporate the FOL loss; the ablated variant is evaluated both with and without it.
 
\begin{itemize}
  \item \textbf{MLP} (2.9\,M parameters):
    A two-layer perceptron applied independently to each clip,
    followed by masked mean pooling.
    This baseline has no temporal reasoning capability and
    serves as the performance floor.
 
  \item \textbf{LSTM} (2.3\,M parameters):
    A bidirectional two-layer LSTM with
    \texttt{pack\_padded\_sequence} for correct handling of
    variable-length sequences.
    This baseline captures long-range sequential dependencies
    without attention.
 
  \item \textbf{Seq-Transformer} (3.07\,M parameters):
    A compact Transformer encoder applied to the fused clip
    sequence, producing predictions at all four levels through
    the same multi-task head design as the proposed model, with
    no graph-structured attention bias.
    This baseline isolates the value of plain self-attention
    over the clip sequence.
 
  \item \textbf{R-GCN} (4.20\,M parameters):
    An external relational graph convolutional network
    \cite{schlichtkrull2018modeling}.
    Each episode is represented as a typed heterogeneous graph whose
    relational edges instantiate the parent relationships of the
    ontology (action to activity, activity to LLI, LLI to HLI), and
    typed message passing is performed over this graph.
    Trained and evaluated under the identical protocol and prediction
    schema, this baseline tests whether an off-the-shelf relational
    architecture can exploit the same structural knowledge.
 
  \item \textbf{Hierarchical Transformer (HT)}
    (3.60\,M parameters):
    The proposed NSGT architecture with the HR-G attention
    bias removed, evaluated both with and without the FOL
    loss.
    This ablation isolates the contribution of graph-structured
    attention from the contribution of the FOL constraint.
\end{itemize}
 
All models produce predictions at all four hierarchy levels and are
trained under the same classification objective as the NSGT conditions
without FOL; the internal baselines share the identical multi-task head module.
Performance differences are therefore attributable to the temporal and
relational reasoning mechanisms rather than to output structure.
 
The primary contribution of this work is not to advance
state-of-the-art action recognition on NTU RGB+D~120, for
which specialised skeleton-based graph convolutional methods
achieve over 90\% top-1 accuracy
\cite{zhou2024blockgcn,do2024skateformer}, but to demonstrate
that a neuro-symbolic graph transformer can jointly reason
across four levels of behavioural abstraction while promoting
ontological consistency.
Action-level results are therefore interpreted in the context
of hierarchical reasoning capability, not as a challenge to
single-level recognition benchmarks.
 
%%─────────────────────────────────────────────────────────────────
\subsection{Evaluation Metrics}
\label{sec:metrics}
%%─────────────────────────────────────────────────────────────────
 
The following metrics are computed on the held-out test set
for all conditions.
 
\textbf{Classification metrics.}
For each of the four hierarchy levels, two metrics are reported:
 
\begin{itemize}
  \item \textbf{Top-1 accuracy}: fraction of correctly
    classified episodes (or clips, for the action level).
 
  \item \textbf{Macro-F1}: unweighted mean of per-class F1
    scores.
    Macro-F1 is selected as the primary classification metric
    because it assigns equal weight to all classes regardless
    of frequency, which is appropriate given the variation in
    episode support across hierarchy levels
    (Table~\ref{tab:dataset}).
 
  \item \textbf{Top-5 accuracy} (action level only): whether
    the ground-truth action class appears among the top-5
    predicted classes.
    This metric is standard in large-scale action recognition
    benchmarks.
 
  \item \textbf{Joint accuracy}: fraction of episodes for which the
    predicted high-level intention is correct and the predicted
    activity set and low-level-intention set both match the
    ground-truth sets exactly.
    This is the strictest measure of hierarchical correctness, since
    a single error at any of the three levels counts as a miss.
\end{itemize}
 
\textbf{Hierarchy consistency rate (HC).}
HC is the primary metric for structural coherence.
For each test episode, hierarchy consistency requires that every predicted activity have at least one of its parent low-level intentions among the predicted low-level intentions, and that every predicted low-level intention have the predicted high-level intention among its parent high-level intentions:
\begin{equation}
  \mathrm{HC} = \frac{1}{N}\sum_{n=1}^{N}
  \mathbb{1}\!\left[
    \hat{\ell}_n \in \mathcal{V}(\hat{g}_n)
    \;\wedge\;
    \hat{h}_n \in \mathcal{V}(\hat{\ell}_n)
  \right]
  \label{eq:lc}
\end{equation}
where $\hat{g}_n$, $\hat{\ell}_n$, $\hat{h}_n$ are the
predicted activity, LLI, and HLI for episode $n$, and
$\mathcal{V}(\cdot)$ denotes the set of valid successors
in the HR-G.
HC is undefined for conditions where not all three levels
are predicted simultaneously, and is reported as N/A in
those cases.
%% ═══════════════════════════════════════════════════════════════════════════════

%% ═══════════════════════════════════════════════════════════════════════════
\section{Results and Analysis}
\label{sec:results}
 
\textbf{Reporting conventions.}
All four hierarchy levels are evaluated consistently under the same
protocol, and the complete per-level metrics are reported together in
Table~\ref{tab:baselines}. Macro-F1 is reported at every level, so that
performance is comparable across levels and robust to class imbalance;
top-1 and top-5 accuracy are additionally reported at the action level,
where a large single-label vocabulary makes top-k accuracy informative.
The hierarchy consistency rate and the joint accuracy summarise
cross-level coherence. The abstract highlights top-1 action accuracy and
the hierarchy consistency rate as headline figures for brevity, but
these are the same quantities reported in full in the results, and no
level is evaluated by a different or less complete procedure than the
others.

\begin{figure*}[t]
  \centering
  \includegraphics[width=\textwidth]{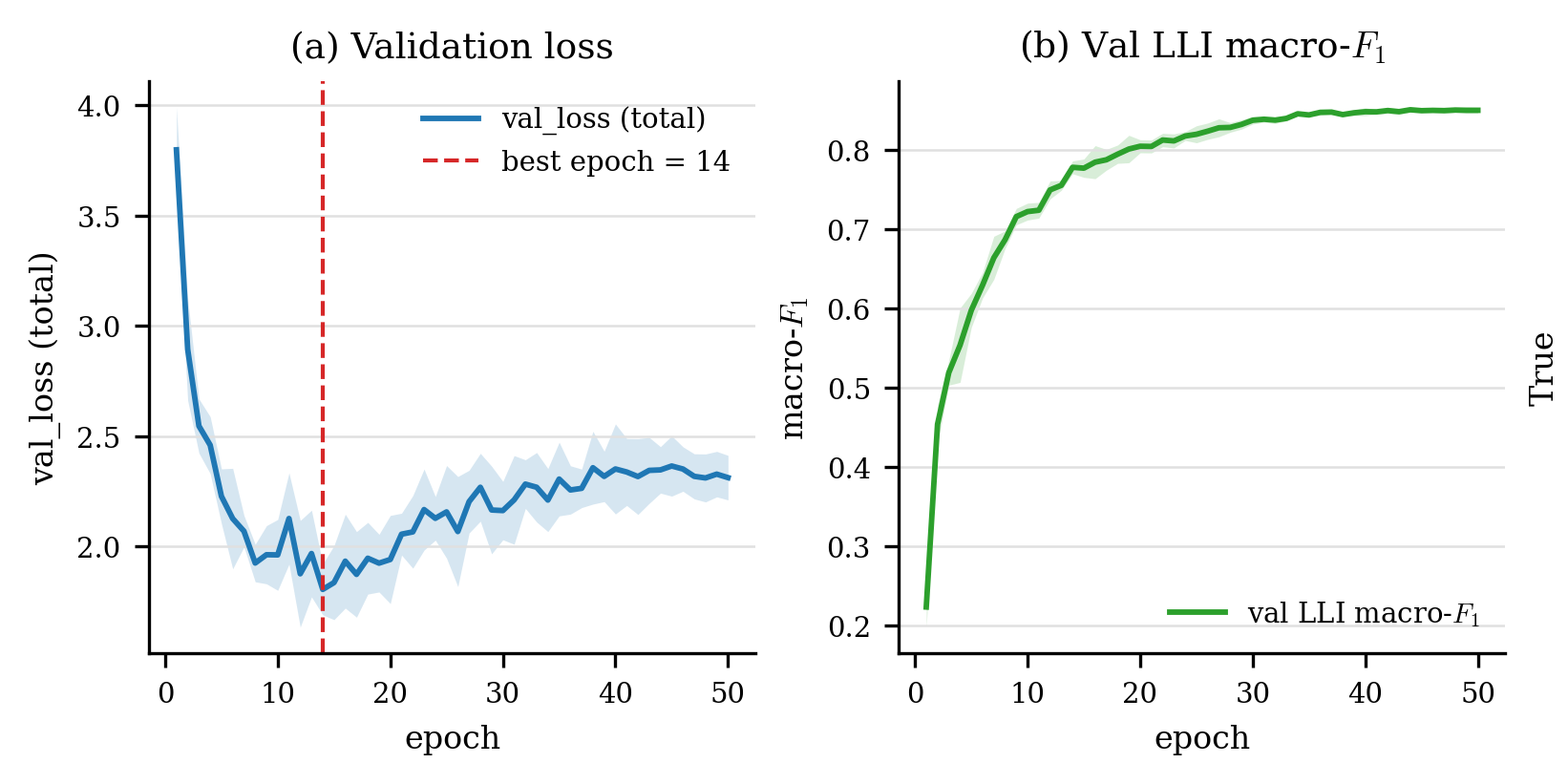}
  \caption{
    Training dynamics of NSGT+FOL ($\lambda=0.5$), averaged over five seeds.
    (a)~Training and validation loss across epochs, with the best
    checkpoint marked at the epoch selected by validation LLI macro-F1.
    The training loss decreases steadily while the validation loss
    flattens and then rises gradually, a calibration effect on the
    softmax heads that does not degrade decision-level performance, as
    the validation macro-F1 in panel~(b) remains stable across the same
    range.
    (b)~Validation LLI macro-F1 across epochs, shown with a
    one-standard-deviation band over seeds.
    LLI macro-F1 is the checkpoint-selection metric.
  }
  \label{fig:training_dynamics}
\end{figure*}
 
Training dynamics of NSGT+FOL ($\lambda=0.5$) are presented in
Figure~\ref{fig:training_dynamics}.
The training loss decreases steadily throughout training.
The validation loss flattens early and then rises gradually; this
reflects growing confidence on the softmax heads rather than a loss
of generalisation, since the validation LLI macro-F1 in
Figure~\ref{fig:training_dynamics}(b) remains stable over the same
range and the reported checkpoint is selected on that metric rather
than on validation loss.
 
%%─────────────────────────────────────────────────────────────────
\subsection{Comparison with Baselines}
\label{sec:baseline_results}
%%─────────────────────────────────────────────────────────────────
 
\begin{figure}[t]
  \centering
  \includegraphics[width=\linewidth]{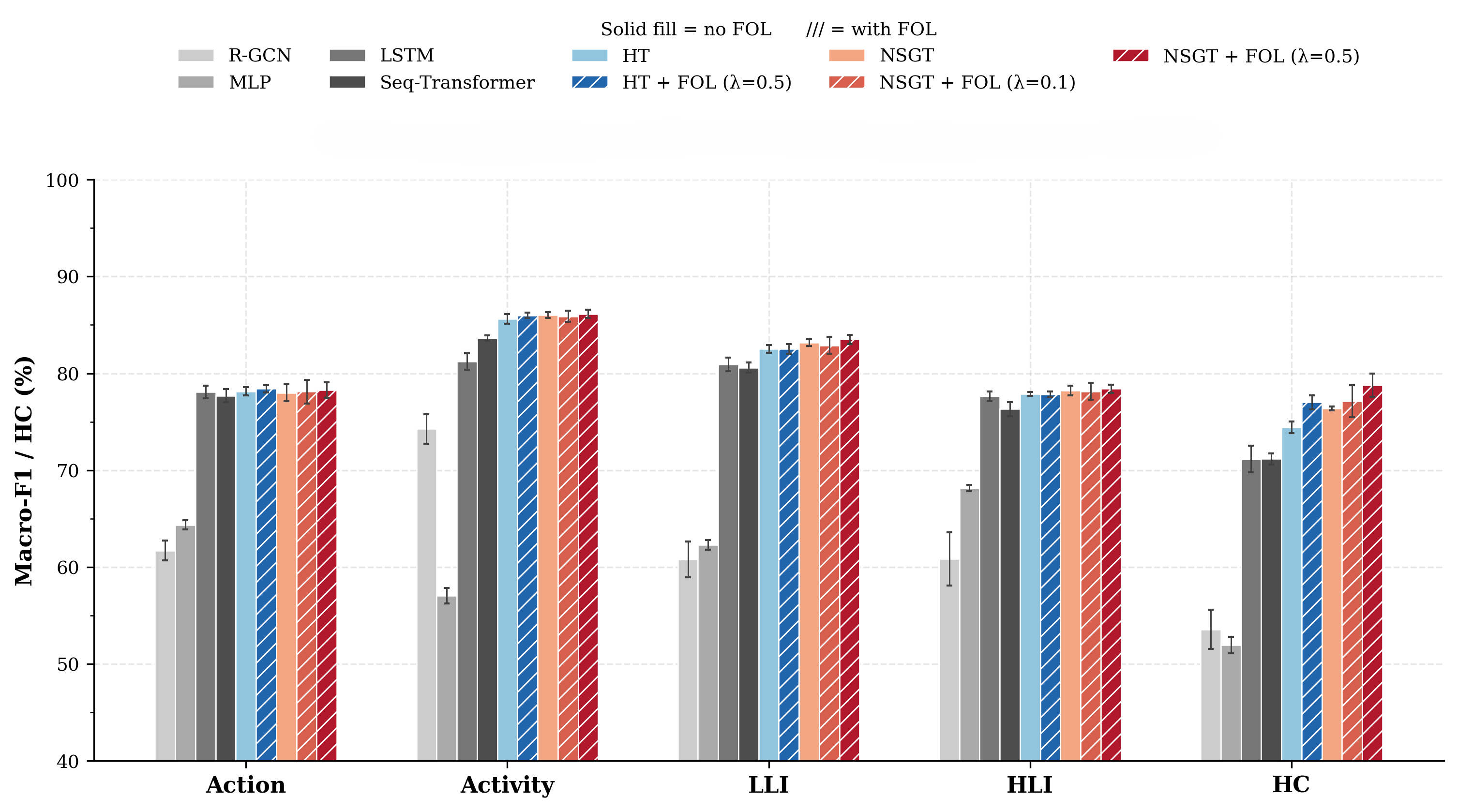}
  \caption{
    Per-level macro-F1 and hierarchy consistency on the held-out test split. Bars report the mean over five seeds and error bars denote one standard deviation. The first four groups report macro-F1 at the action, activity, low-level intention, and high-level intention levels, computed over all classes of the taxonomy (Table~\ref{tab:taxonomy}). The fifth group reports the hierarchy consistency rate, which measures the proportion of predictions satisfying the parent-child constraints of the ontology and is therefore not a macro-F1 score. Hatched bars denote conditions trained with the first-order logic objective. Improvements attributable to the logic objective are concentrated in hierarchy consistency rather than in per-level macro-F1. The vertical axis starts at 40\% to resolve differences among the stronger conditions.
  }
  \label{fig:bar_baselines}
\end{figure}

Full numerical results are reported in Table~\ref{tab:baselines}.
Figure~\ref{fig:bar_baselines} provides a visual comparison of all evaluated
conditions across the four hierarchy levels and the hierarchy consistency rate.
Two patterns are visible. First, the improvement attributable to the first-order
logic objective is concentrated in hierarchy consistency rather than in per-level
discrimination. For the hierarchical transformer, hierarchy consistency is raised
from 74.42\% to 77.02\% by the logic objective, while per-level macro-F1 remains
within 0.4 points at every level. The same behaviour is observed for the proposed
model, where hierarchy consistency is raised from 76.40\% to 78.79\% and
per-level macro-F1 varies by at most 0.4 points. This separation indicates that the
logic objective operates as a constraint on the joint prediction structure rather
than as an additional source of discriminative signal at any individual level.
Second, at a matched logic setting the proposed model attains higher hierarchy
consistency than the hierarchical transformer, 78.79\% against 77.02\%, and the
same ordering is observed without the logic objective, 76.40\% against 74.42\%,
which indicates that the architectural and symbolic contributions are additive.
The unstructured and graph baselines remain well below all transformer variants on
hierarchy consistency, at 51.94\% and 53.56\% respectively, despite the graph
baseline reaching 74.26\% macro-F1 at the activity level, which shows that
competitive performance at a single level does not imply coherent predictions across
the hierarchy.
 
\begin{tableorg}[t]
\centering
\caption{
  Classification performance and hierarchy consistency on the
  NTU-Hierarchy test set.
  Act-T1: action top-1 accuracy (\%).
  Act-F1: action macro-F1 (\%).
  Atvty-F1: activity macro-F1 (\%).
  HC: hierarchy consistency rate (\%). Joint: joint exact-match accuracy (\%).
  Mean $\pm$ standard deviation over five seeds. Best mean per column in \textbf{bold}.
}
\label{tab:baselines}
\scriptsize
\setlength{\tabcolsep}{2pt}
\begin{adjustbox}{max width=\textwidth}
\begin{tabular}{lrrrrrrrr}
\toprule
\textbf{Model} &
\textbf{Act-T1} & \textbf{Act-T5} & \textbf{Act-F1} &
\textbf{Atvty-F1} & \textbf{LLI-F1} & \textbf{HLI-F1} &
\textbf{HC} & \textbf{Joint} \\
\midrule
MLP  & 75.48\sd{0.89} & 96.62\sd{0.23} & 64.35\sd{0.49} & 57.05\sd{0.82} & 62.28\sd{0.49} & 68.16\sd{0.31} & 51.94\sd{0.86} & 17.02\sd{0.24} \\
LSTM & 86.89\sd{0.37} & 98.46\sd{0.17} & 78.07\sd{0.66} & 81.24\sd{0.85} & 80.91\sd{0.70} & 77.63\sd{0.49} & 71.16\sd{1.40} & 29.07\sd{0.79} \\
Seq-Transformer & 86.53\sd{0.35} & 98.92\sd{0.04} & 77.70\sd{0.67} & 83.64\sd{0.27} & 80.59\sd{0.53} & 76.32\sd{0.71} & 71.18\sd{0.57} & 31.92\sd{0.45} \\
R-GCN           & 77.32\sd{0.64} & 96.97\sd{0.29} & 61.70\sd{1.03} & 74.26\sd{1.54} & 60.79\sd{1.84} & 60.85\sd{2.73} & 53.56\sd{2.02} & 17.70\sd{1.28} \\
\midrule
HT (no FOL)  & 86.96\sd{0.29} & 98.98\sd{0.09} & 78.15\sd{0.42} & 85.64\sd{0.50} & 82.55\sd{0.40} & 77.86\sd{0.20} & 74.42\sd{0.61} & 35.43\sd{0.48} \\
HT + FOL     & 87.02\sd{0.26} & 98.98\sd{0.12} & \textbf{78.41}\sd{0.37} & 85.99\sd{0.28} & 82.52\sd{0.50} & 77.84\sd{0.29} & 77.02\sd{0.73} & 35.82\sd{0.78} \\
\midrule
NSGT (no FOL) &
  87.32\sd{0.58} & 98.89\sd{0.07} & 78.00\sd{0.88} & 86.01\sd{0.29} & 83.18\sd{0.34} & 78.24\sd{0.48} & 76.40\sd{0.20} & 36.39\sd{0.46} \\
NSGT + FOL ($\lambda$=0.1) &
  87.21\sd{0.71} & \textbf{98.99}\sd{0.10} & 78.12\sd{1.24} & 85.89\sd{0.58} & 82.88\sd{0.87} & 78.14\sd{0.87} & 77.15\sd{1.64} & 35.98\sd{2.27} \\
NSGT + FOL ($\lambda$=0.5) &
  \textbf{87.37}\sd{0.44} & 98.98\sd{0.16} & 78.30\sd{0.80} & \textbf{86.14}\sd{0.43} & \textbf{83.51}\sd{0.47} & \textbf{78.41}\sd{0.43} & \textbf{78.79}\sd{1.19} & \textbf{36.90}\sd{0.74} \\
\bottomrule
\end{tabular}
\end{adjustbox}
\end{tableorg}

\paragraph{Overall ranking.}
At the action level the transformer models are closely matched,
and NSGT+FOL ($\lambda$=0.5) is comparable to the strongest
sequential baseline (LSTM) on action macro-F1 (78.30\% vs.\
78.07\%). The advantage of the proposed model concentrates at the
higher levels of the hierarchy and in structural coherence.
Relative to LSTM, NSGT+FOL improves activity macro-F1 by
4.90\,pp (86.14\% vs.\ 81.24\%), LLI macro-F1 by 2.60\,pp
(83.51\% vs.\ 80.91\%), and HLI macro-F1 by 0.78\,pp (78.41\%
vs.\ 77.63\%), while hierarchy consistency improves by 7.63\,pp
(78.79\% vs.\ 71.16\%). The same ordering holds against the
hierarchical transformer without the graph bias, confirming that
the graph-structured attention and the logic objective contribute
gains in hierarchical consistency rather than in single-level
action discrimination.
The attention-based sequence transformer tracks LSTM closely,
while the external relational-graph baseline trails at every
level despite its explicit relational structure (61.70\% action
F1, 53.56\% hierarchy consistency), which indicates that generic
message passing over the ontology does not substitute for the
proposed ontology-biased attention.
 
\paragraph{Interpretation of action-level results.}
The action-level results of HT and NSGT are not directly
comparable with specialised skeleton-based recognition methods
such as ST-GCN~\cite{yan2018spatial} or
CTR-GCN~\cite{chen2021channel}, which achieve over 90\%
top-1 accuracy on NTU RGB+D~120 using architectures
specifically designed for joint-topology graphs.
The models in this work are optimised for joint hierarchical
prediction across four levels simultaneously.
The action head therefore operates under a multi-task
constraint, and some action-level accuracy is foregone in
exchange for joint prediction across all four hierarchy levels.
This is by design: the primary objective is
intention recognition, not isolated action classification.
 
\paragraph{Contextual positioning.}
To provide broader context, NSGT is compared with selected
recent multimodal and multi-task frameworks evaluated on
NTU RGB+D~120 in Table~\ref{tab:sota_comparison}.
These methods operate on flat action labels rather than
four-level hierarchical annotations, so the comparison is
indicative rather than equivalent.
 
\begin{table}[t]
\centering
\caption{
  Indicative comparison with recent multimodal and multi-task
  frameworks on NTU RGB+D~120.
  Comparison models predict flat action labels; NSGT jointly
  predicts four hierarchy levels.
  Results are not directly comparable due to differences in
  task formulation.
}
\label{tab:sota_comparison}
\small
\begin{tabularx}{\columnwidth}{@{} l l X r @{}}
\toprule
\textbf{Method} & \textbf{Year} &
\textbf{Strategy} & \textbf{Acc (\%)} \\
\midrule
MMINet \cite{cheng2022spatial}   & 2022 &
  Cross-modality interactive learning & 91.7 \\
MMNet \cite{bruce2022mmnet}      & 2023 &
  Model-level attention transfer & 92.4 \\
M2-CLIP \cite{wang2024multimodal}& 2024 &
  Semantic alignment and task adapters & 89.2 \\
BHaRNet \cite{cho2025body}       & 2025 &
  Body-hand joint recognition (MoE) & 93.0 \\
Spathis et al.\ \cite{spathis2025multi} & 2025 &
  Joint action and gesture (SPS) & 88.5 \\
\midrule
NSGT (no FOL) & 2025 &
  Hierarchical graph transformer & 87.32 \\
NSGT + FOL ($\lambda$=0.5) & 2025 &
  Hierarchical + logic regularisation & 87.37 \\
\bottomrule
\end{tabularx}
\end{table}
 
At the action level, NSGT attains 87.37\% top-1 accuracy while
additionally producing four levels of hierarchical prediction
and a quantified hierarchy consistency rate.
The remaining gap to specialised flat-label systems reflects
the multi-task constraint discussed above, and the logic
objective imposes no accuracy cost, with 87.32\% top-1 without
and 87.37\% with the FOL loss.
Structural coherence across a four-level hierarchy is a
property that flat-label methods do not model.

%%─────────────────────────────────────────────────────────────────
\subsection{Lookup-Table Baseline}
\label{sec:lookup}
%%─────────────────────────────────────────────────────────────────
 
To test whether the higher levels of the hierarchy can be recovered
from action predictions alone, a deterministic lookup-table baseline is
evaluated. This baseline takes the model's predicted actions and maps
them to the activity, low-level intention, and high-level intention
levels through the ontology, applying no learned reasoning above the
action level.
 
The lookup is fully specified as follows. The predicted action of every real clip slot in an episode is mapped to its parent activities in the ontology; the activity prediction of the episode is the union of these parents, restricted to the evaluation taxonomy, so that several valid parents are all retained rather than resolved by choice. The low-level intention prediction is the union of the parents of the predicted activities, restricted in the same way. The high-level intention prediction is the single class receiving the largest number of predicted low-level intentions among its children; ties are broken deterministically in favour of the lowest class index. Variable episode length is absorbed by the union over clip slots, so no per-episode length normalisation is required. The only inputs are the ontology tables and the model's own action predictions; no training-set statistics, test labels or test-derived class frequencies are used at any step.
 
\begin{table}[t]
\centering
\caption{
  Full model versus the deterministic lookup-table baseline
  (macro-F1, mean $\pm$ std over 5 seeds, NSGT + FOL $\lambda$=0.5).
  The lookup maps predicted actions to the higher levels through the
  ontology. A positive gap indicates the model outperforms the lookup.
}
\label{tab:lookup}
\small
\begin{tabular}{llrrr}
\toprule
\textbf{Split} & \textbf{Level} & \textbf{Model} & \textbf{Lookup} & \textbf{Gap} \\
\midrule
\multirow{3}{*}{Test}
 & Activity & $86.14 \pm 0.43$ & $87.58 \pm 0.45$ & $-1.44$ \\
 & LLI      & $83.51 \pm 0.47$ & $45.33 \pm 0.29$ & $+38.18$ \\
 & HLI      & $78.41 \pm 0.43$ & $27.88 \pm 0.36$ & $+50.53$ \\
\midrule
\multirow{3}{*}{Compositional}
 & Activity & $67.11 \pm 0.54$ & $69.99 \pm 0.31$ & $-2.88$ \\
 & LLI      & $59.21 \pm 0.62$ & $36.40 \pm 0.08$ & $+22.81$ \\
 & HLI      & $47.02 \pm 0.57$ & $25.57 \pm 0.13$ & $+21.45$ \\
\bottomrule
\end{tabular}
\end{table}
 
At the activity level, the lookup baseline is competitive and slightly
exceeds the model, by 1.44 pp on the standard test split. This is
expected, because an activity is composed of one to three actions
and is therefore close to a deterministic function of the predicted
actions; where the mapping is nearly one-to-one, a lookup is an
appropriate and strong baseline. At the intention levels, however, the
picture reverses sharply. The full model exceeds the lookup baseline by
38.18 pp at the low-level intention level and by 50.53 pp at the
high-level intention level on the standard test split, and the lookup
collapses toward chance at the high-level intention level (27.88\%
macro-F1). The one-to-many design of the ontology means that a given set
of actions may be consistent with several intentions, so the lookup
requires an explicit resolution rule. The same pattern holds on
the compositional test split, where the model exceeds the lookup
baseline by 22.81 pp at the low-level intention level and 21.45 pp at
the high-level intention level. These results show that the reported
lookup implementation does not recover the intention-level performance
of the learned model; they do not establish that every possible
non-neural inference rule would fail.
 
%%─────────────────────────────────────────────────────────────────
\subsection{Generalisation to Unseen Compositions}
\label{sec:comp_results}
 
Beyond the standard subject-disjoint test split, performance is
evaluated on the compositional test set, which withholds from training one high-level parent association of each multi-parent low-level intention. This split measures whether the
learned reasoning transfers to novel arrangements of familiar actions
rather than memorised co-occurrence patterns.
 
\begin{tableorg}[t]
\centering
\caption{
  Performance on the compositional test split (macro-F1 unless noted,
  mean $\pm$ standard deviation over five seeds). Act: action. Atvty: activity. LLI/HLI: low- and
  high-level intention. HC: hierarchy consistency rate. Joint: joint
  accuracy across all four levels. Best result per column in
  \textbf{bold}.
}
\label{tab:comp}
\scriptsize
\setlength{\tabcolsep}{3pt}
\begin{adjustbox}{max width=\textwidth}
\begin{tabular}{lrrrrrr}
\toprule
\textbf{Model} & \textbf{Act} & \textbf{Atvty} & \textbf{LLI} & \textbf{HLI} & \textbf{HC} & \textbf{Joint} \\
\midrule
MLP                       & 58.89\sd{0.38} & 42.19\sd{0.63} & 42.66\sd{0.81} & 44.11\sd{0.51} & 38.60\sd{1.50} & 3.63\sd{0.40} \\
LSTM                      & \textbf{64.32}\sd{0.62} & 62.82\sd{1.10} & 59.54\sd{0.33} & \textbf{49.26}\sd{0.72} & 64.11\sd{1.68} & 9.07\sd{0.88} \\
Seq-Transformer           & 63.63\sd{0.26} & 65.09\sd{0.58} & 56.67\sd{1.14} & 46.42\sd{0.71} & 63.20\sd{1.39} & 11.07\sd{0.77} \\
R-GCN                     & 52.78\sd{1.46} & 56.54\sd{1.76} & 40.00\sd{1.64} & 36.09\sd{2.20} & 40.01\sd{1.99} & 0.54\sd{0.28} \\
\midrule
HT                        & 63.33\sd{0.27} & 66.62\sd{0.21} & 57.98\sd{0.89} & 46.19\sd{0.36} & 68.77\sd{0.92} & 12.98\sd{0.62} \\
HT + FOL ($\lambda$=0.5)  & 63.59\sd{0.27} & 66.87\sd{0.40} & 58.47\sd{0.79} & 46.65\sd{0.64} & 70.40\sd{0.96} & 13.57\sd{0.98} \\
NSGT                      & 63.40\sd{0.62} & 66.99\sd{0.69} & 59.16\sd{0.87} & 47.03\sd{0.77} & 68.84\sd{0.90} & 12.54\sd{0.71} \\
NSGT + FOL ($\lambda$=0.1)& 63.27\sd{0.90} & 66.75\sd{1.13} & 58.86\sd{1.26} & 47.47\sd{1.25} & 69.41\sd{2.47} & 13.23\sd{1.42} \\
NSGT + FOL ($\lambda$=0.5)& 63.49\sd{0.44} & \textbf{67.11}\sd{0.54} & \textbf{59.21}\sd{0.62} & 47.02\sd{0.57} & \textbf{72.15}\sd{1.06} & \textbf{13.63}\sd{0.53} \\
\bottomrule
\end{tabular}
\end{adjustbox}
\end{tableorg}

Overall performance is lower than on the standard test split, as
expected when generalising to unseen compositions. The full model,
NSGT + FOL at $\lambda=0.5$, attains the highest hierarchy consistency
rate (72.15\%) and the highest joint accuracy across all four levels
(13.63\%), exceeding the strongest competing configuration on both
measures. The relational-graph baseline transfers poorly, with a joint
accuracy near zero, which indicates that its representation does not
recover coherent multi-level structure under distribution shift.
 
The effect of the logic objective is more pronounced on this split
than on the standard test set. On the standard test split, adding the
logic objective raises the hierarchy consistency rate of the full
model from 76.40\% to 78.79\%, a gain of 2.39 pp. On the compositional
split, the same objective raises it from 68.84\% to 72.15\%, a gain of
3.31 pp. This larger gain is consistent with a stronger regularization
effect under compositional shift. Because the metric is aligned with
the ontology-derived loss, it should not be interpreted by itself as
independent evidence of general reasoning.

%%─────────────────────────────────────────────────────────────────
\subsection{Ablation Study: The Role of Joint Training}
\label{sec:ablation_joint}
%%─────────────────────────────────────────────────────────────────
 
Table~\ref{tab:head_ablation} reports the head ablation on the
HT backbone, isolating the contribution of each output level.
 
\begin{table}[t]
\centering
\caption{
  Head ablation on the HT backbone, averaged over the three
  ablation-campaign seeds.
  Conditions (1)--(4): subset of heads active.
  Conditions (5)--(6): all four heads with and without FOL;
  these are the full HT conditions of Table~\ref{tab:baselines}
  and agree with it within seed variation.
  HC: hierarchy consistency, undefined (N/A) for partial-head
  conditions.
  Dashes mark heads excluded from the loss in that
  condition; their untrained outputs are not reported.
}
\label{tab:head_ablation}
\small
\begin{tabular}{lrrrrr}
\toprule
\textbf{Condition} &
\textbf{Act-F1} & \textbf{Atvty-F1} & \textbf{LLI-F1} &
\textbf{HLI-F1} & \textbf{HC} \\
\midrule
(1) Action only              & 78.16 & --- & --- & --- & N/A \\
(2) Activity only            & --- & 79.27 & --- & --- & N/A \\
(3) Intention only (LLI+HLI) & --- & --- & 68.30 & 67.89 & N/A \\
(4) Action + Activity        & 79.33 & 86.18 & --- & --- & N/A \\
\midrule
(5) Full HT (no FOL)         & 78.00 & 85.29 & 82.76 & 77.84 & 74.64 \\
(6) Full HT + FOL            & 78.21 & 85.86 & 82.70 & 77.80 & \textbf{77.26} \\
\bottomrule
\end{tabular}
\end{table}
 
\paragraph{Joint training benefits the intention levels.}
Comparisons are drawn only between heads that are
actively supervised in each condition. Supervising the two intention
levels alone (C3) attains 68.30\% low-level and 67.89\% high-level
intention F1; adding the lower-level heads under the full objective
(C5) raises these to 82.76\% and 77.84\%. The improvement at the most
abstract level shows that lower-level supervision shapes a
representation the intention heads exploit, rather than the levels
being independent. That the intention levels are not recoverable from
lower-level outputs is established separately, and without appeal to
untrained heads, by the deterministic lookup baseline
(Section~\ref{sec:lookup}): a given action or activity composition is
consistent with several distinct intentions, so the lower levels
underdetermine the intention.
 
\paragraph{No action-level cost from joint training.}
Action-only supervision (C1, 78.16\%) and the full four-head
model (C5, 78.00\%) attain essentially the same action F1,
within seed variation.
Supervising the three additional levels therefore adds
hierarchical capability at no measurable cost to action-level
discrimination, which indicates that the shared encoder has
sufficient capacity for the joint objective.
 
\paragraph{FOL effect on hierarchy consistency.}
The FOL loss improves hierarchy consistency by approximately
2.6 points on the HT backbone, consistent with
Table~\ref{tab:fol_ablation}.
Classification differences between C5 and C6 remain within
0.6\,pp across all levels, confirming that the FOL loss acts as
a structural regulariser rather than a classification booster.
 
%%─────────────────────────────────────────────────────────────────
\subsection{Error Structure under Partial Supervision}
\label{sec:head_ablation}
%%─────────────────────────────────────────────────────────────────
 
\begin{figure}[t]
  \centering
  \begin{subfigure}[t]{0.48\textwidth}
    \includegraphics[width=\textwidth]{%
      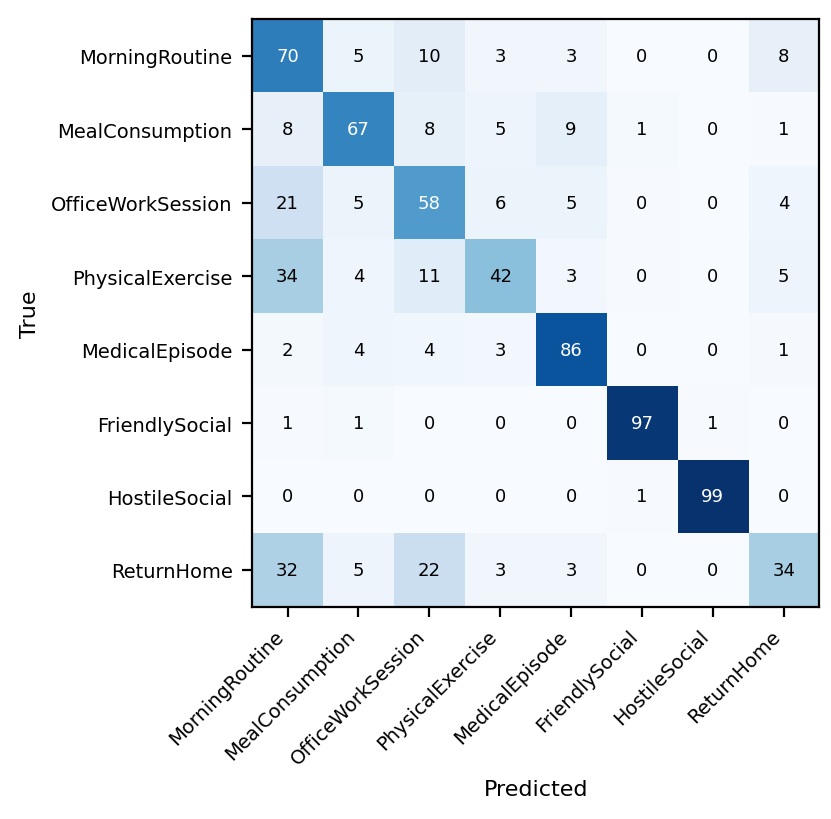}
    \caption{Intention only (recall: 69.2\%)}
  \end{subfigure}
  \hfill
  \begin{subfigure}[t]{0.48\textwidth}
    \includegraphics[width=\textwidth]{%
      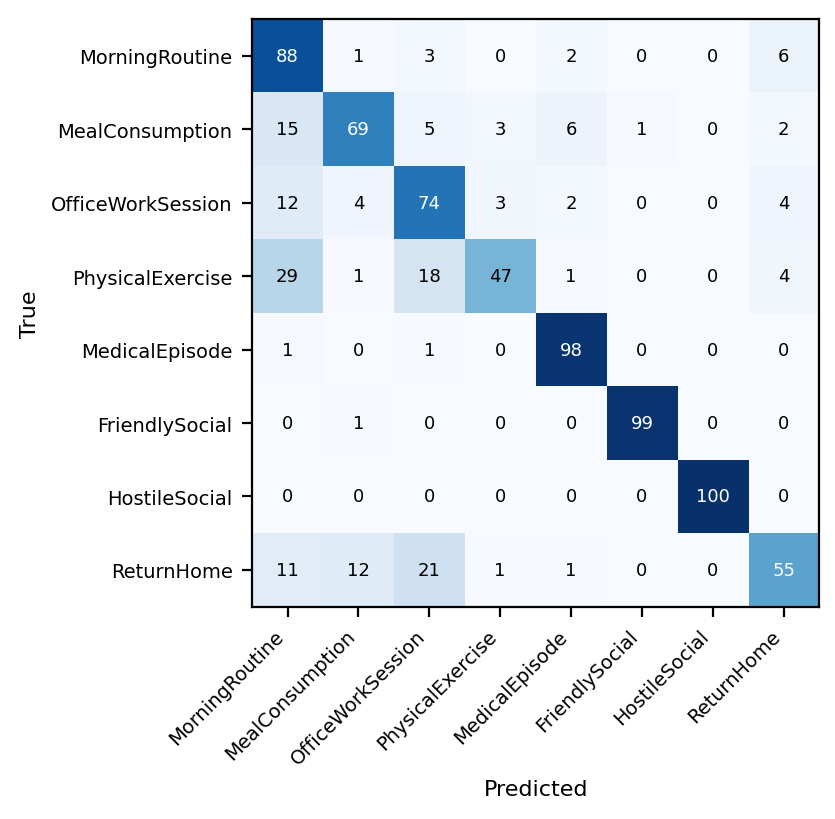}
    \caption{Full objective (recall: 78.7\%)}
  \end{subfigure}
  \caption{High-level intention confusion matrices under two
supervision configurations in which the high-level head is actively supervised,
evaluated on the held-out test split at seed 42 and normalised by true-class row.
Cells report the percentage of episodes of each true class predicted as each
candidate class. The left panel is supervised at the two intention levels only, the
right panel under the full hierarchical objective. Chance recall is 12.5 percent.
Under intention-only supervision the mean diagonal recall is 69.2 percent, and under
the full objective it reaches 78.7 percent, with residual confusions concentrated
among ontologically adjacent classes. The improvement shows that the lower-level
heads contribute non-redundant information to the highest level under joint
training.}
  \label{fig:cm_head_ablation}
\end{figure}

The contribution of the lower-level heads to the highest level is
isolated by comparing two configurations in which the high-level intention head is
actively supervised (Figure~\ref{fig:cm_head_ablation}). When supervision is provided
at the two intention levels alone, the mean diagonal recall for high-level intention
is 69.2 percent, and the confusions that remain concentrate among semantically
adjacent classes, such as the return-home class being confused with the
morning-routine and office classes. Adding the lower-level heads under the full
objective raises the mean diagonal recall to 78.7 percent, with the residual errors
concentrating further on ontologically adjacent classes that share transitional
sub-intentions, as analysed in Appendix~\ref{secA1}. The intention levels are
therefore improved by the presence of the lower-level heads during joint training,
which indicates that the levels carry non-redundant information coupled by the
hierarchical objective. That the lower levels alone cannot recover high-level
intention is established by the lookup-table baseline of Section~\ref{sec:lookup},
which quantifies the one-to-many structure directly, rather than from the behaviour
of an unsupervised head.
 
The two directions of this result are complementary: the lower levels alone cannot recover high-level intention, as established by the lookup-table baseline, and the intention levels are in turn improved by the presence of the lower-level heads during joint training, so the levels carry non-redundant information that the hierarchical objective couples.
 
%%─────────────────────────────────────────────────────────────────
\subsection{Impact of Graph-Structured Attention}
\label{sec:graph_impact}
%%─────────────────────────────────────────────────────────────────
 
\paragraph{Where the graph bias contributes.}
At a matched logic setting, the graph-structured attention
leaves action-level discrimination essentially unchanged
relative to HT+FOL (78.30\% vs.\ 78.41\% action F1) and
instead contributes at the intention levels and in structural
coherence: +0.99\,pp LLI F1, +0.57\,pp HLI F1, and +1.77\,pp
hierarchy consistency.
The same pattern holds without the logic objective, where the
graph bias raises hierarchy consistency by +1.98\,pp over the
uniform-attention backbone (76.40\% vs.\ 74.42\%) and LLI F1
by +0.63\,pp.
This distribution of gains is consistent with the learned bias
weights: the dominant low-level-intention co-membership edge
links clips that share an intention, a relation that
constrains the joint prediction across levels rather than the
identity of any single action.
At the LLI and HLI levels, the label distribution is
determined by episode-level ontological structure, which
mean-pooling in the HT already captures reasonably well.
 
\paragraph{Analysis of learned $\beta$ parameters.}
Table~\ref{tab:betas} reports the converged values of
the three HR-G bias scalars across all NSGT conditions.
 
\begin{table}[t]
\centering
\caption{
  Converged HR-G bias weights at the end of training, reported as the mean and standard
  deviation over five seeds. Larger values indicate stronger exploitation of the
  corresponding edge type. The low-level-intention co-membership edge is dominant across
  all three NSGT conditions. The random-graph control row uses the three-seed ablation protocol.
}
\label{tab:betas}
\small
\begin{tabular}{lrrr}
\toprule
\textbf{Condition} &
$\boldsymbol{\beta}_{\mathrm{trans}}$ &
$\boldsymbol{\beta}_{\mathrm{act}}$ &
$\boldsymbol{\beta}_{\mathrm{lli}}$ \\
\midrule
NSGT (no FOL)              & $0.093 \pm 0.022$ & $0.069 \pm 0.017$ & $0.601 \pm 0.011$ \\
NSGT + FOL ($\lambda$=0.5) & $0.091 \pm 0.020$ & $0.067 \pm 0.018$ & $0.603 \pm 0.016$ \\
NSGT + FOL ($\lambda$=0.1) & $0.092 \pm 0.015$ & $0.070 \pm 0.014$ & $0.597 \pm 0.028$ \\
NSGT, random graph (no FOL) & $0.020 \pm 0.101$ & $0.026 \pm 0.065$ & $0.132 \pm 0.055$ \\
\bottomrule
\end{tabular}
\end{table}
 
All conditions converge to the same qualitative pattern:
$\beta_{\mathrm{lli}} \approx 0.60$ is dominant, while
$\beta_{\mathrm{trans}} \approx 0.09$ and
$\beta_{\mathrm{act}} \approx 0.07$ remain small and positive.
The low-level-intention co-membership edge is therefore the most strongly exploited,
directing attention between clips that share a low-level intention, while temporal
transition and same-activity edges contribute a smaller positive bias. The ranking is
stable across all five seeds, with cross-seed standard deviations below 0.03 for every
scalar.
 
This pattern is interpretable within the predefined HR-G: under the tested
configuration, low-level-intention co-membership receives the largest learned
coefficient. This does not establish that the graph topology itself is optimal.
 
\begin{figure}[t]
  \centering
  \includegraphics[width=\linewidth]{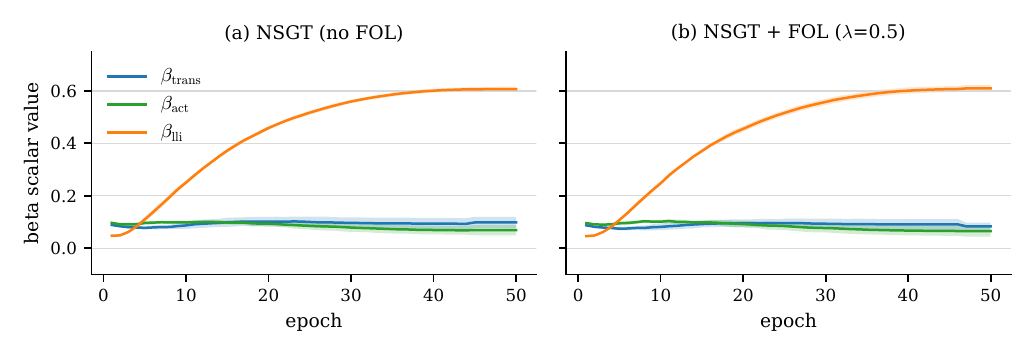}
  \caption{Evolution of the three HR-G bias scalars during training, averaged over five seeds, with shaded bands denoting one standard deviation. Panel (a): trained without the logic objective; panel (b): trained with the logic objective at $\lambda=0.5$. In both settings $\beta_{\mathrm{lli}}$, initialised at the smallest value of the three, overtakes the other two scalars within the first epochs and continues to grow throughout training, reaching approximately 0.61 at the final epoch, while $\beta_{\mathrm{trans}}$ and $\beta_{\mathrm{act}}$ remain near their initialisation. The two panels are nearly indistinguishable, indicating that the logic objective does not change which edge type the attention exploits.}
  \label{fig:betas}
\end{figure}

Figure~\ref{fig:betas} shows the evolution of the three HR-G bias scalars over training, with and without the logic objective. The scalar ordering is established within the first few epochs and remains stable: $\beta_{\mathrm{lli}}$, initialised at the smallest value, overtakes $\beta_{\mathrm{trans}}$ and $\beta_{\mathrm{act}}$, while the latter two remain close to their initialisation. The two panels are also similar, suggesting that the tested logic weight has little effect on the relative coefficient ordering. These trajectories demonstrate parameter stability, not robustness to a different graph topology.
 
Validation LLI macro-F1 is used for model selection because LLI is the
primary intention-level target and macro-F1 limits domination by frequent
classes. The same criterion is applied to all compared models. This is a
design choice rather than a theoretically unique criterion, and any
single-level checkpoint criterion may favor that level.

\textbf{Matched-density random-graph control.} To separate the contribution of the graph's content from that of its density, the three relation tables are replaced by random graphs with identical off-diagonal edge counts (215 directed successor edges; 44 and 251 undirected co-membership pairs), zero diagonal, and support restricted to the evaluation taxonomy, and the model is retrained under the three-seed ablation protocol without the logic loss. Two effects are observed. First, the learned bias scalars collapse by an order of magnitude and lose their cross-seed stability, with the low-level-intention co-membership weight falling from $0.601 \pm 0.011$ on the correct graph to $0.132 \pm 0.055$ on the random one (Table~\ref{tab:betas}): the training signal suppresses a topology that carries no ontological information. Second, performance degrades gracefully rather than collapsing: hierarchy consistency reaches $75.93 \pm 1.12$ and joint accuracy $35.93 \pm 0.49$, between the no-graph transformer ($74.42$, $35.43$) and the correct graph ($76.40$, $36.39$), with the residual differences from either neighbour within seed variation at three seeds. The graph is therefore a content-sensitive inductive bias rather than a tuned auxiliary: the correct topology is exploited with strong, stable weights, whereas an uninformative topology of equal density is downweighted towards zero and neither reproduces the correct graph's contribution nor harms the model below the graph-free baseline.

%%─────────────────────────────────────────────────────────────────
\subsection{FOL Loss Analysis}
\label{sec:fol_analysis}
 
\textbf{Definition and measurement of contradictions.}
A prediction is termed ontologically contradictory when the labels it
assigns across the four levels cannot coexist under the ontology.
The hierarchy consistency rate reported throughout is the fraction of test episodes whose predicted labels form a valid chain through the ontology: every predicted activity is required to have at least one of its parent low-level intentions among the predicted low-level intentions, and every predicted low-level intention is required to have the predicted high-level intention among its parent high-level intentions (equation~\eqref{eq:lc}). Activity mutual-exclusion violations enter the training loss as a separate Type-C penalty but are not included in this aggregate evaluation rate. The reported metric therefore covers the two parent--child conditions only.
 
\textbf{Behaviour under ambiguous or conflicting evidence.}
The value of a soft constraint is most apparent where perceptual
evidence is ambiguous. Because each violation term is continuous in the
predicted probabilities, the penalty it applies is graded: when the
evidence is uncertain, the violation changes smoothly with the
probabilities rather than imposing a hard threshold. A diffuse
prediction does not necessarily imply a small penalty for every rule. When
strong perceptual evidence conflicts with an ontological expectation,
the penalty grows with the degree of violation but does not override the classification signal, because it is weighted by $\lambda$ against the supervised loss. The practical consequence is visible in the
results: increasing the logic weight raises the hierarchy-consistency
rate monotonically while leaving action accuracy nearly unchanged. This
demonstrates empirical alignment under the tested settings; it does not
identify which individual predictions were corrected because their
perceptual evidence was weak.
 
%%─────────────────────────────────────────────────────────────────
 
Table~\ref{tab:fol_ablation} reports the effect of the FOL
weight $\lambda_{\mathrm{FOL}}$ on both backbones.
 
\begin{table}[t]
\centering
\caption{
  FOL loss ablation on the HT and NSGT backbones.
  $\Delta$HC: hierarchy consistency improvement over the no-FOL
  baseline of the same backbone.
}
\label{tab:fol_ablation}
\small
\setlength{\tabcolsep}{3.5pt}
\begin{tabular}{llrrrrrr}
\toprule
\textbf{Backbone} & $\boldsymbol{\lambda}$ &
\textbf{Act-F1} & \textbf{Atvty-F1} &
\textbf{LLI-F1} & \textbf{HLI-F1} & \textbf{HC} &
$\Delta$\textbf{HC} \\
\midrule
HT   & ---  & 78.15 & 85.64 & 82.55 & 77.86 & 74.42 & --- \\
HT   & 0.5  & 78.41 & 85.99 & 82.52 & 77.84 & 77.02 & +2.60 \\
\midrule
NSGT & ---  & 78.00 & 86.01 & 83.18 & 78.24 & 76.40 & --- \\
NSGT & 0.1  & 78.12 & 85.89 & 82.88 & 78.14 & 77.15 & +0.75 \\
NSGT & 0.5  & 78.30 & 86.14 & 83.51 & 78.41 & \textbf{78.79} & +2.39 \\
\bottomrule
\end{tabular}
\end{table}
 
The FOL loss produces a measurable and consistent improvement
in hierarchy consistency at negligible cost to classification
accuracy.
At $\lambda$=0.1, action F1 changes by only 0.12\,pp relative
to the no-FOL baseline (78.12\% vs.\ 78.00\%) while hierarchy
consistency improves by +0.75\,pp.
At $\lambda$=0.5, hierarchy consistency improves by +2.39\,pp, while per-level classification metrics change by less than 0.4\,pp at every level and are unchanged within seed variation.
 
This behaviour has practical implications for deployment.
$\lambda$=0.1 already provides a meaningful consistency gain
with essentially no change in discriminative performance.
In deployment settings such as assistive robotics, where ontological coherence is a first-order requirement, $\lambda$=0.5 provides the largest reduction in logical contradictions among the configurations tested, at no cost to classification accuracy. This improvement in consistency at no cost to accuracy is a practical advantage of the proposed framework over purely data-driven approaches.
 
%%─────────────────────────────────────────────────────────────────
\subsection{Modality Ablation}
\label{sec:modality_results}
%%─────────────────────────────────────────────────────────────────
 
Table~\ref{tab:modality} reports the effect of restricting
the input to a single sensor modality.
NSGT+FOL ($\lambda$=0.5) is retrained from scratch for each
configuration, with the input projection layer dimension
adjusted accordingly.
 
\begin{table}[t]
\centering
\caption{
  Modality ablation: NSGT+FOL ($\lambda$=0.5) trained on
  unimodal versus full four-modality input, averaged over
  three seeds. The input projection dimension adjusts to the
  active modalities (2048 for skeleton, 1024 for RGB, 5120 for
  all four). All other settings are identical.
}
\label{tab:modality}
\small
\setlength{\tabcolsep}{4pt}
\begin{tabular}{lrrrrrr}
\toprule
\textbf{Modality} &
\textbf{Act-T1} & \textbf{Act-F1} &
\textbf{Atvty-F1} & \textbf{LLI-F1} & \textbf{HLI-F1} &
\textbf{HC} \\
\midrule
All modalities &
  \textbf{87.37} & \textbf{78.30} &
  \textbf{86.14} & \textbf{83.51} & \textbf{78.41} &
  \textbf{78.79} \\
Skeleton only &
  55.41 & 39.75 & 63.95 & 59.74 & 60.66 & 64.09 \\
RGB only &
  79.60 & 70.43 & 79.36 & 77.28 & 74.49 & 71.90 \\
Infrared only &
  68.36 & 60.08 & 74.26 &
  71.61 & 70.26 & 70.82 \\
Depth only &
  74.37 & 67.00 & 77.90 &
  75.21 & 72.81 & 70.80 \\
\bottomrule
\end{tabular}
\end{table}
 
\begin{figure}[t]
  \centering
  \begin{subfigure}[t]{0.32\textwidth}
    \includegraphics[width=\textwidth]{%
      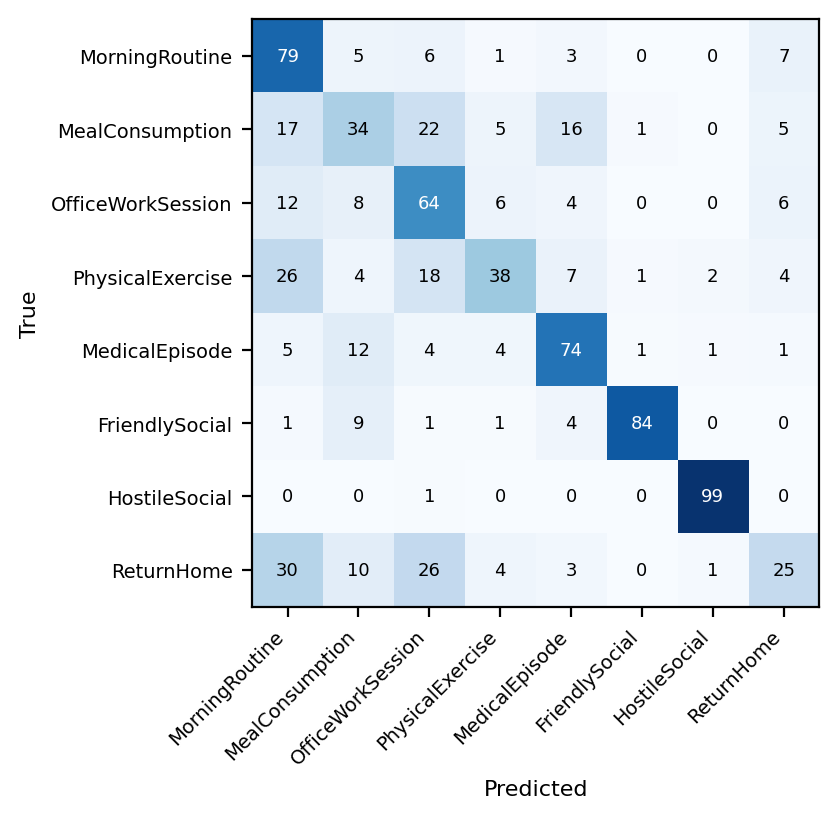}
    \caption{Skeleton only (HLI-F1: 60.7\%)}
  \end{subfigure}
  \hfill
  \begin{subfigure}[t]{0.32\textwidth}
    \includegraphics[width=\textwidth]{%
      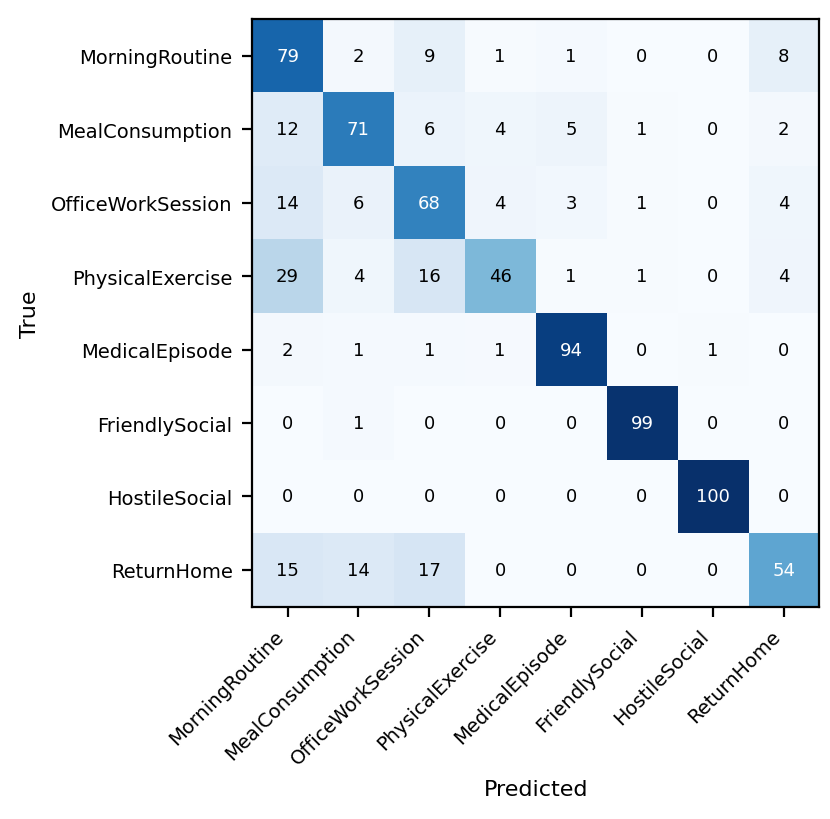}
    \caption{RGB only (HLI-F1: 74.5\%)}
  \end{subfigure}
  \hfill
  \begin{subfigure}[t]{0.32\textwidth}
    \includegraphics[width=\textwidth]{%
      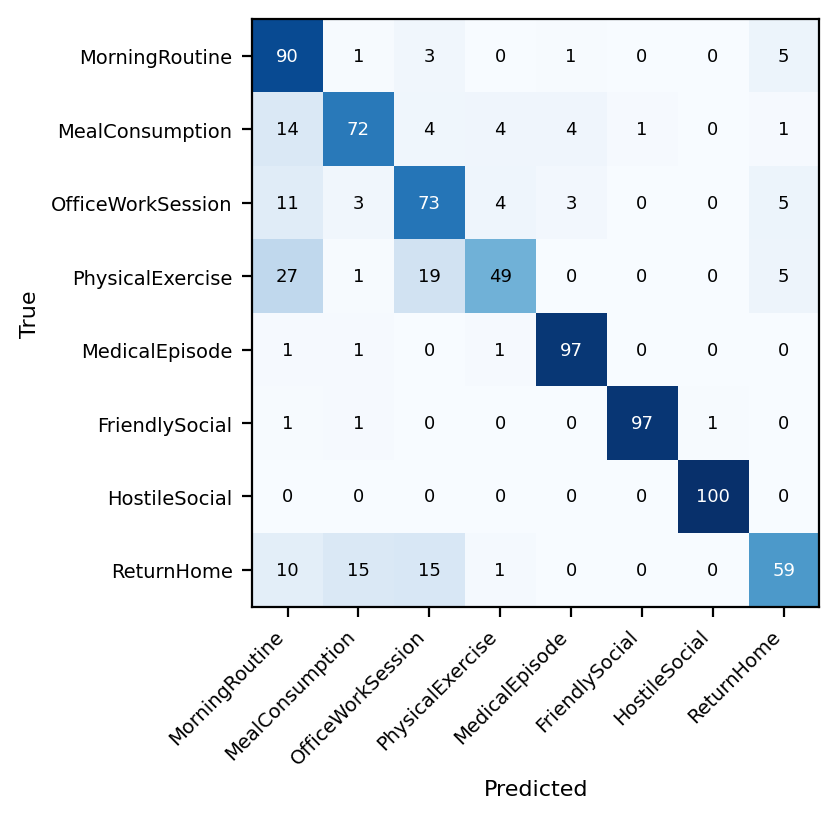}
    \caption{All modalities (HLI-F1: 78.4\%)}
  \end{subfigure}
  \caption{
    Row-normalised high-level-intention confusion matrices
    (test set, values in \%) for the three modality conditions.
    Under skeleton-only input, the context-dependent classes
    collapse: meal consumption is recognised at 33.5\% and is
    confused with office work (22\%), and return-home is
    recognised at 25.2\% and confused with morning routine
    (30\%) and office work (26\%), since these episode types are
    distinguished by scene and objects rather than by body pose.
    RGB restores both classes (meal consumption 70.7\%,
    return-home 53.6\%, medical episode 94.2\%), and full fusion
    improves them further (meal consumption 71.5\%, return-home
    59.0\%). Residual errors shared by all three conditions, such
    as physical exercise confused with morning routine, persist
    regardless of modality because these classes share transitional
    sub-intentions and are therefore ontologically overlapping rather
    than modality-limited (Appendix~\ref{secA1}).
  }
  \label{fig:cm_modality}
\end{figure}

\paragraph{Skeleton as the primary kinematic signal.}
Restricting the input to skeleton reduces action F1 by
38.55\,pp (78.30\% $\to$ 39.75\%) and hierarchy consistency by
14.70\,pp (78.79\% $\to$ 64.09\%). The PoseC3D embeddings encode
body-joint kinematics, which remain discriminative for classes
defined by limb configuration, but without appearance context
the model cannot resolve episode types that depend on the
objects and scene present. This is visible at the high-level
intention level, where meal consumption and return-home fall to
33.5\% and 25.2\% recall respectively (Figure~\ref{fig:cm_modality}). These are the intention
classes that are composed largely of transitional sub-intentions shared
across several high-level intentions, so their separation rests on scene
and object evidence rather than on the shared body-motion pattern; the
ontological basis of this overlap is analysed in
Appendix~\ref{secA1}.
 
\paragraph{RGB as a complementary appearance signal.}
RGB-only reaches 70.43\% action F1 and 79.36\% activity F1,
confirming the discriminative strength of the VideoMAE-v2
features. The 7.87\,pp action-F1 gap to the full model indicates
that RGB captures complementary scene and object information
that skeleton does not encode; the two context-dependent classes
that collapse under skeleton, meal consumption and return-home,
recover to 70.7\% and 53.6\% under RGB.

\textbf{Infrared and depth streams.} The two remaining unimodal conditions complete the picture: depth-only reaches 74.37\% action top-1 and infrared-only 68.36\%, both between the skeleton and RGB extremes, and their hierarchy consistency (70.80\% and 70.82\%) remains well below the fused model's 78.79\%. No single stream dominates the fused representation: the strongest unimodal configuration (RGB) trails the full model by 7.77 pp in action top-1 and 3.92 pp in high-level intention macro-F1, and each unimodal condition loses between 6.9 and 14.7 pp of hierarchy consistency, which indicates that cross-level coherence draws on complementary evidence from several streams.

\paragraph{Non-redundant fusion.}
The full four-modality model outperforms every unimodal condition at every hierarchy level. No single stream dominates the fusion, and the two context-dependent classes recovered by RGB are further supported by the additional streams. Hierarchy
consistency degrades most in the skeleton-only condition
(64.09\%), which indicates that appearance features support
ontological coherence beyond kinematic discrimination.
 
Figure~\ref{fig:cm_modality} provides qualitative evidence for
the complementarity claim. Skeleton-only errors are concentrated
in the intention classes that depend on scene and object context,
which RGB is able to resolve. Full modality fusion attains the
highest recall on these classes while the residual errors, which
are shared across conditions, correspond to semantically adjacent
episode types.

%% ═══════════════════════════════════════════════════════════════════════════
\section{Discussion and Limitations}
\label{sec:discussion}
 
\subsection{Consistency Gains at No Accuracy Cost}
\label{sec:tradeoff}
 
A central finding is that the logic objective improves the reported hierarchy-consistency diagnostic while per-level classification metrics remain nearly unchanged. On the NSGT backbone the consistency gain at $\lambda_{\mathrm{FOL}}=0.5$ is statistically significant across the five seeds (paired $t$-test, $p=0.009$), whereas the accompanying changes in per-level macro-F1 are within seed variation and none reaches significance. The same separation holds on the HT backbone. This shows that the regularizer changes cross-level coherence more strongly than per-level discrimination under the tested protocol.
 
The hierarchy-consistency rate is a discrete chain check, whereas the
training term uses continuous probabilities and also includes Type-C
constraints. Nevertheless, the metric and loss are ontology-aligned and
are not independent. The gain is therefore interpreted as an
objective-aligned diagnostic. Complementary evidence comes from
per-level metrics, the graph-only comparison without FOL, the lookup
baseline, and the compositional split.
 
On the compositional split, the consistency gain from the logic objective
is +3.31\,pp for NSGT and +1.63\,pp for HT, compared with +2.39\,pp and
+2.60\,pp respectively on the standard test split. This difference is
consistent with an interaction between graph-biased attention and the
logic term under compositional shift, but the aggregate ablation does
not by itself identify the mechanism responsible for that interaction.
 
Among the tested values, $\lambda_{\mathrm{FOL}}=0.5$ gives the highest
hierarchy consistency with no statistically significant reduction in
per-level macro-F1. This recommendation is limited to the present
benchmark and evaluation protocol.
 
%%─────────────────────────────────────────────────────────────────
\subsection{Progressive Architectural Gains}
\label{sec:progressive}
%%─────────────────────────────────────────────────────────────────
 
The results in Table~\ref{tab:baselines} reveal a progressive
performance improvement that mirrors the increasing architectural
complexity of the evaluated models.
 
MLP achieves 64.35\% action F1 with no temporal reasoning.
This establishes that temporal context is essential for the
benchmark: features from isolated clips, without any sequential
modeling, are insufficient for reliable action recognition.
 
LSTM and the HT backbone demonstrate the benefit of sequential
and attention-based temporal reasoning respectively, with action
F1 rising to 78.07\% and 78.15\%.
At the action level the sequential, attention, and graph models
are closely matched, which indicates that action recognition on
sequences of mean length 7.04 clips (7.36 on the compositional
split) is largely saturated by temporal modelling alone. The sequence Transformer tracks the LSTM closely at every level, which indicates that at this episode scale attention and recurrence recover the same temporal information. The relational-graph baseline, by contrast, trails the multilayer perceptron at the action, low-level, and high-level intention levels despite its explicit ontological structure, and its hierarchy consistency is comparable to that of the perceptron; structural knowledge alone, without fine-grained temporal modelling, does not yield coherent predictions, a mechanism examined in Appendix~\ref{secA1}.
 
The graph structure and the logic objective instead deliver their
gains at the higher levels of the hierarchy and in structural
coherence.
NSGT+FOL ($\lambda$=0.5) improves hierarchy consistency to
78.79\%, a +7.63 point improvement over LSTM and +1.77 points
over HT+FOL, together with the best activity, LLI, and HLI
macro-F1 in the suite.
This confirms that encoding ontological relationships directly into the attention mechanism provides structural information beyond what uniform attention over raw features can recover.
 
Finally, the FOL loss contributes a consistent and reproducible
improvement in hierarchy consistency across both backbones:
+2.60 points for HT and +2.39 points for NSGT.
Classification differences with and without FOL remain within 0.4 points at every level, confirming that the FOL loss acts as
a structural regulariser rather than a classification booster.
 
This progression, from temporal modeling to global attention
to graph-structured attention to logic regularisation, supports
the paper's central claim: each architectural component addresses
a specific limitation of the previous one, and the full NSGT+FOL
system is the product of principled, incremental design.
 
%%─────────────────────────────────────────────────────────────────
\subsection{Interpretability and the Audit Trail}
\label{sec:audit}
%%─────────────────────────────────────────────────────────────────
 
\begin{figure}[!htbp]
  \centering
  \begin{subfigure}[t]{0.48\textwidth}
    \centering
    \includegraphics[width=1.1\linewidth]{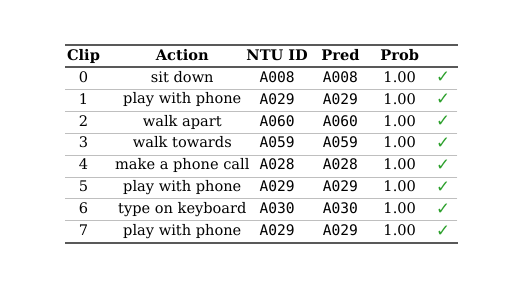}
    \caption{Clips, ground truth and second-pass predictions; all eight correct with probability above 0.99.}
    \label{fig:audit_a}
  \end{subfigure}\hfill
  \begin{subfigure}[t]{0.48\textwidth}
    \centering
    \includegraphics[width=0.8\linewidth]{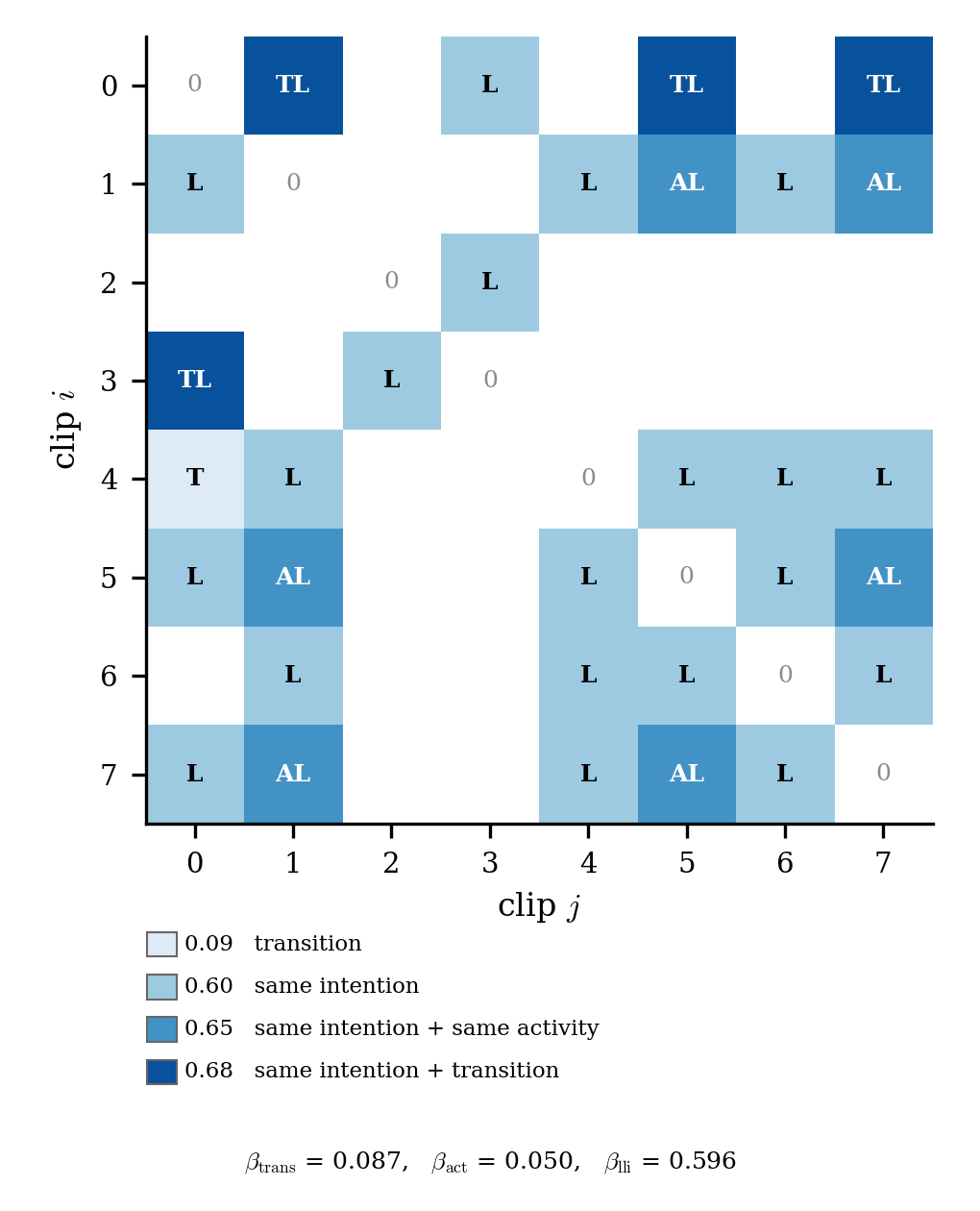}
    \caption{HR-G bias matrix built from the first-pass predicted actions.}
    \label{fig:audit_b}
  \end{subfigure}
  \\[1ex]
  \begin{subfigure}[t]{0.48\textwidth}
    \centering
    \includegraphics[width=\linewidth]{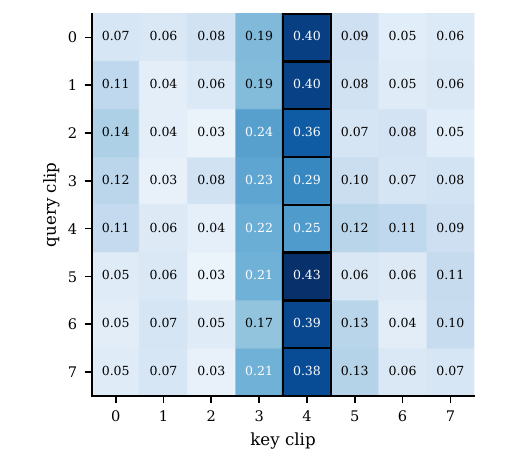}
    \caption{Final-layer attention, averaged over heads.}
    \label{fig:audit_c}
  \end{subfigure}\hfill
  \begin{subfigure}[t]{0.48\textwidth}
    \centering
    \includegraphics[width=0.9\linewidth]{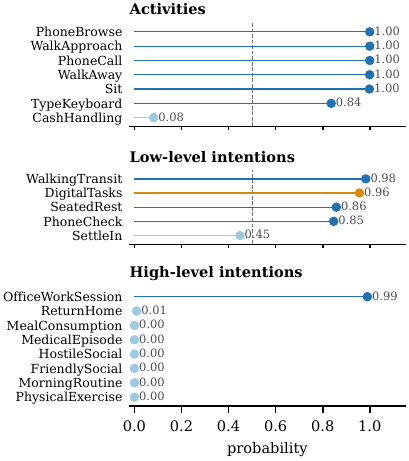}
    \caption{Predicted probabilities over the evaluation taxonomy.}
    \label{fig:audit_d}
  \end{subfigure}
  \vspace{2mm}
  \caption{
    Audit trail of held-out test episode \texttt{ep\_0004228} (seed-42 checkpoint, NSGT with $\lambda=0.5$). Thirty of the 31 nonzero entries of the bias matrix carry the same-intention relation, weighted by the largest of the three learned scalars ($\beta_{\mathrm{trans}}=0.09$, $\beta_{\mathrm{act}}=0.05$, $\beta_{\mathrm{lli}}=0.60$). The resulting parent-intention resolution is shown in Figure~\ref{fig:audit_resolution}.
  }
  \label{fig:audit}
\end{figure}

\paragraph{A worked audit-trail example.}
Figure~\ref{fig:audit} traces one held-out test episode, \texttt{ep\_0004228}, of eight clips through the seed-42 checkpoint of NSGT with $\lambda=0.5$, from input to verdict. The recorded clips are, in order, sitting down, browsing a phone, walking apart, walking towards, making a phone call, browsing a phone, typing on a keyboard and browsing a phone; both passes predict all eight actions correctly with probabilities above 0.99. The six activities recovered from them (\texttt{Sit}, \texttt{PhoneBrowse}, \texttt{WalkAway}, \texttt{WalkApproach}, \texttt{PhoneCall}, \texttt{TypeKeyboard}) are mostly ambiguous in the ontology: \texttt{Sit} alone has seven candidate parent intentions and each walking activity four, whereas \texttt{TypeKeyboard} has exactly one, \texttt{DigitalTasks}. The bias matrix $B$ built from the first-pass actions (Figure~\ref{fig:audit_b}) has 31 nonzero entries, 30 of which carry the same-intention co-membership relation weighted by $\beta_{\mathrm{lli}} = 0.60$, so attention is drawn towards clips that share a candidate intention; in the final layer every clip attends most strongly to the phone-call clip and the walk-towards clip (Figure~\ref{fig:audit_c}). Four low-level intentions pass the 0.5 threshold with clear margins (Figure~\ref{fig:audit_d}) (\texttt{WalkingTransit} 0.98, \texttt{DigitalTasks} 0.96, \texttt{SeatedRest} 0.86, \texttt{PhoneCheck} 0.85), and \texttt{SettleIn} is rejected at 0.45. Three of the four are transitional intentions that belong to four high-level intentions each, and the only high-level intention other than \texttt{OfficeWorkSession} compatible with all three is \texttt{ReturnHome}; \texttt{DigitalTasks}, whose sole parent is \texttt{OfficeWorkSession}, resolves the ambiguity, and the head assigns 0.99 to \texttt{OfficeWorkSession} and 0.01 to \texttt{ReturnHome}, with every other class below $10^{-3}$. The chain check passes at both parent conditions and the prediction is a joint exact match. A reviewer can therefore read from the trace which clips drove the decision, which goals were competing, and which single piece of evidence separated the two plausible ones.
 
\begin{figure}[!htbp]
  \centering
  \resizebox{0.90\textwidth}{!}{%
    \begin{tikzpicture}[
      every node/.style={font=\scriptsize},
      lli/.style={draw, rounded corners=2pt, fill=blue!8, minimum width=22mm, minimum height=5.5mm},
      decide/.style={draw=orange!80!black, very thick, rounded corners=2pt, fill=orange!12, minimum width=22mm, minimum height=5.5mm},
      hli/.style={draw=gray!55, rounded corners=2pt, fill=gray!4, minimum width=26mm, minimum height=5.5mm},
      sel/.style={draw=green!45!black, very thick, rounded corners=2pt, fill=green!8, minimum width=26mm, minimum height=5.5mm},
      amb/.style={draw=gray!70, rounded corners=2pt, fill=gray!8, minimum width=26mm, minimum height=5.5mm},
      e/.style={draw=gray!45, line width=0.35pt},
      ed/.style={draw=orange!80!black, line width=1.1pt}
    ]
      \node[decide] (dt) at (0,3.0) {DigitalTasks};
      \node[lli]    (pc) at (0,2.0) {PhoneCheck};
      \node[lli]    (wt) at (0,1.0) {WalkingTransit};
      \node[lli]    (sr) at (0,0.0) {SeatedRest};
      \node[sel] (ow) at (5.4,3.0) {OfficeWorkSession};
      \node[amb] (rh) at (5.4,2.0) {ReturnHome};
      \node[hli] (mr) at (5.4,1.0) {MorningRoutine};
      \node[hli] (mc) at (5.4,0.0) {MealConsumption};
      \node[hli] (pe) at (5.4,-1.0) {PhysicalExercise};
      \draw[e] (pc) -- (mr); \draw[e] (pc) -- (mc); \draw[e] (pc) -- (ow); \draw[e] (pc) -- (rh);
      \draw[e] (wt) -- (mr); \draw[e] (wt) -- (ow); \draw[e] (wt) -- (pe); \draw[e] (wt) -- (rh);
      \draw[e] (sr) -- (mc); \draw[e] (sr) -- (ow); \draw[e] (sr) -- (pe); \draw[e] (sr) -- (rh);
      \draw[ed] (dt) -- (ow);
      \node[anchor=north west, text=gray!60!black] at (-1.4,-1.9)
        {Three transitional intentions leave \{OfficeWorkSession, ReturnHome\} plausible;};
      \node[anchor=north west, text=orange!70!black] at (-1.4,-2.35)
        {DigitalTasks has a single parent and decides.};
    \end{tikzpicture}}
  \caption{Parent-intention resolution for the held-out episode of Figure~\ref{fig:audit}. Candidate parent intentions of each predicted low-level intention. Three transitional intentions leave \texttt{OfficeWorkSession} and \texttt{ReturnHome} plausible; \texttt{DigitalTasks}, whose only parent is \texttt{OfficeWorkSession}, resolves the ambiguity. The chain check passes at both parent conditions and the prediction is a joint exact match.}
  \label{fig:audit_resolution}
\end{figure}
 
%%─────────────────────────────────────────────────────────────────
\subsection{Training Convergence}
\label{sec:convergence}
%%─────────────────────────────────────────────────────────────────
 
The training curves in Figure~\ref{fig:training_dynamics}
show a consistent pattern.
The training loss decreases smoothly throughout training,
indicating stable optimisation without gradient instability.
From the middle of training onward the validation loss rises
gradually while the training loss continues to decrease; this is a
calibration effect on the softmax heads rather than a collapse in
accuracy, since the validation macro-F1 remains stable over the same
range.
Model selection uses the epoch of highest validation LLI macro-F1,
which is insensitive to this calibration drift, so the reported
checkpoint retains the highest decision-level performance.

%%─────────────────────────────────────────────────────────────────
\subsection{The HLI Ceiling}
\label{sec:limit_hli}
%%─────────────────────────────────────────────────────────────────
 
The LSTM and all transformer variants reach 76--78\% HLI macro-F1, against a chance level of 12.5\% for eight classes. The narrow spread across architectures indicates that the residual errors at this level are not architectural, and their structure shows what they are instead.
 
\paragraph{Semantic overlap as a deliberate benchmark property.}
The ontology preserves one-to-many relationships by design: four transitional low-level intentions (walking transit, seated rest, phone check, and drink break) are each valid constituents of four different high-level intentions, so their presence in an episode cannot by itself determine the intention. This overlap partitions the eight classes into groups with different empirical behaviour. Classes composed of shared transitional intentions absorb much of the residual error, whereas several classes with more distinctive constituents are recognised more accurately. Cross-model confusion matrices and the skeleton/RGB modality comparison support this interpretation empirically. They do not constitute an ontology-independent validation, because identifying the shared constituents still relies on the designed ontology. Semantic overlap is therefore presented as a plausible explanation of the observed confusion pattern rather than as a proven source of the performance ceiling.
 
Two directions follow from this diagnosis. First, longer temporal context would supply the disambiguating evidence that a single activity-level episode lacks: a return-home episode and a morning routine share transitional behaviour but are distinguished by what precedes them, so observing sequences of episodes rather than one episode of mean length 7.04 clips (7.36 on the compositional split) would test whether the HR-G transition structure gains discriminative value at longer horizons, beyond its currently modest contribution relative to the dominant low-level-intention edge. Second, explicit scene-context priors, such as time-of-day or static environmental and object anchors, would inject precisely the evidence on which the shared classes are separable, and could be introduced as an additional conditioning input without altering the reasoning architecture.
 
%%─────────────────────────────────────────────────────────────────
\subsection{Synthesized Hierarchical Labels}
\label{sec:limit_labels}
%%─────────────────────────────────────────────────────────────────
 
The four-level hierarchical labels are synthesized from the
NTU RGB+D~120 action classes using the proposed ontology,
rather than collected from human annotators.
This was a deliberate and necessary methodological choice:
no multimodal hierarchical intention dataset with the breadth
of NTU RGB+D~120 existed at the time of this work.
The annotation pipeline, transition graph, and episode
synthesis procedure were developed as a research contribution
in their own right.
 
The inherent limitation is that the synthesized labels encode the
ontological assumptions of the designers. The FOL constraints are also
derived from that ontology, so they are not an independent source of
semantic validation. The deterministic lookup baseline
(Section~\ref{sec:lookup}) falls 38.18\,pp and 50.53\,pp short of the
model at the two intention levels, showing that the reported lookup
implementation does not recover the model's performance. Outputs from
heads excluded from the training loss are not treated as valid evidence,
because those heads are not trained predictors.
 
A second limitation concerns ecological validity. The episodes are assembled from staged laboratory recordings of single performers, and their compositions are curated by the generator rather than observed in the wild. The benchmark therefore tests whether a model can recover hierarchical structure from realistic perceptual evidence under controlled compositional and performer shift, which is the property the reasoning components are designed to supply; it does not test naturalistic long-horizon behaviour, with its interruptions, concurrent goals, and uncontrolled scene context. Both the subject-consistency constraint and the transition-governed ordering are intended to keep the assembled episodes behaviourally coherent, but they are approximations of continuity rather than recordings of it.
 
Validation on datasets with independently human-annotated
intention labels remains an important future step.
Crowdsourced annotation or LLM-assisted labeling of NTU
RGB+D~120 clips at the activity and intention levels would
provide ground-truth labels independent of any specific
ontology.
 
%%─────────────────────────────────────────────────────────────────
\subsection{Graph Architecture and Temporal Scale}
\label{sec:limit_graph}
%%─────────────────────────────────────────────────────────────────
 
The current NSGT applies ontology-structured additive attention biases rather than propagating node features along HR-G edges through typed neighbourhood operations, so the graph operates as an attention modulation on top of a temporal backbone rather than as the primary representational substrate. The external relational-graph baseline shows why this arrangement matters: typed message passing over the same ontological edges, without a temporal sequence model beneath it, reaches a hierarchy consistency of 53.56\% and collapses at the high-level intention level, because static aggregation over shared sub-actions cannot break the ties that the transitional classes create (Appendix~\ref{secA1}). Relational structure is therefore valuable only when it is coupled to fine-grained temporal modelling. A variant in which typed message passing between activity, LLI, and HLI nodes is layered on the temporal Transformer, rather than substituted for it, is the architectural direction that this result supports.
 
A related constraint is the temporal scale of the experiments.
Episodes are constructed at the activity level, mean 7.04 clips
(7.36 on the compositional split), with a structural minimum of
four clips.
At this scale, the transition edge receives a smaller weight because successor
adjacency is sparse at the action level, whereas the low-level-intention
co-membership edge, which links clips across the multiple activities that make up
an episode, is the most informative relational signal, consistent with the
dominant $\beta_{\mathrm{lli}}$ reported above.
The HR-G transition structure shows its full benefit when applied
to longer sequences spanning multiple activities and LLIs.
Extending the framework to HLI-level episode sequences of
10--20 clips is therefore identified as the primary
architectural direction for future investigation.
 
The FOL rules in the current framework encourage hierarchical
ontological consistency
(Activity $\to$ LLI $\to$ HLI co-occurrence) but do not
directly constrain temporal ordering within episodes.
Temporal ordering is delegated to the Transformer's positional
encodings and the asymmetric $\beta_{\mathrm{trans}}$ bias.
The introduction of temporal logic constraints, for example
rules of the form
$\textit{Before}(A_i, A_j) \Rightarrow \textit{Activity}(X, B)$,
would make the temporal dimension of reasoning explicit and
verifiable, and is proposed as a direction for a subsequent
study.
 
%%─────────────────────────────────────────────────────────────────
\subsection{Multimodal Fusion Strategy}
\label{sec:limit_modality}
%%─────────────────────────────────────────────────────────────────
 
The current framework employs early fusion: the four modality
feature vectors are concatenated into a single 5120-dimensional
representation before projection into the Transformer's
embedding space.
While computationally efficient, this design merges all modality
signals at the input layer, preventing the model from attending
across modalities independently.
A cross-modal attention architecture, in which separate token
sequences are maintained for each modality and attention is
computed both within and across modality streams, would allow
the model to identify explicit correspondences between, for
example, skeleton joint configurations and RGB appearance
regions.
 
The current modality ablation (Section~\ref{sec:modality_results})
shows that full fusion outperforms skeleton-only and RGB-only input.
Infrared-only and depth-only results are needed to attribute the gain
to each stream separately. The observed fusion gain suggests that a
late-fusion or cross-modal architecture
could recover additional performance currently lost in the
early-fusion projection.
In real-world deployment, skeleton extraction may additionally
be unreliable under occlusion, at long range, or in crowded
scenes.
An uncertainty-aware fusion strategy that down-weights
unreliable modalities at inference time would provide more
robust behaviour in assistive contexts.
 
%%═════════════════════════════════════════════════════════════════
\section{Conclusion and Future Work}
\label{sec:conclusion}
%%═════════════════════════════════════════════════════════════════
 
A neuro-symbolic framework for hierarchical human intention
recognition has been presented. The Neuro-Symbolic Graph Transformer
(NSGT) jointly predicts atomic actions, composite activities,
Low-Level Intentions, and High-Level Intentions from four sensor
modalities, organised in a four-level ontology whose relational
structure is encoded as a Human Reasoning Graph. Three contributions
are evaluated through controlled experiments on a synthesized hierarchical benchmark derived from NTU RGB+D~120. The principal comparison uses five seeds, while the secondary head and modality ablations use three.
 
First, the ontology-structured graph attention, implemented as three
learned HR-G bias scalars, improves hierarchy consistency by +1.98
points over the uniform-attention backbone without the logic objective
and by +1.77 points with it, and yields the best intention-level
macro-F1 in the evaluated suite, while action-level discrimination
remains comparable across the transformer models. The reported
deterministic lookup is exceeded by 38.18 and 50.53 points at the two
intention levels, and an external relational-graph baseline
that performs typed message passing over the same ontology falls well
below the proposed model, so the effective coupling is ontology-biased
attention over a temporal backbone rather than relational structure
alone. The converged $\beta$ values provide interpretable evidence of
which edge types the model finds informative: the low-level-intention
co-membership edge is the most strongly exploited
($\beta_{\mathrm{lli}} \approx 0.60$) under the tested configuration.
The proposed audit-trail structure will be illustrated with a real
held-out episode so that the activated edges, attention values, outputs,
and chain check can be inspected together.
 
Second, the differentiable FOL loss provides a consistent and tunable
improvement in hierarchy consistency across both the HT (+2.60 points)
and NSGT (+2.39 points) backbones, significant across seeds, without
requiring hard IF-THEN rules that would be invisible to
backpropagation. Because per-level accuracy is unchanged within seed
variation at every setting, $\lambda_{\mathrm{FOL}}$ selects the
strength of the consistency gain rather than a trade against accuracy.
On the compositional split, where tested combinations are unseen in
training, the full model attains the highest hierarchy consistency and
joint accuracy of the suite, and the consistency gain from the logic
objective grows under this shift for the proposed model while it
shrinks for the uniform-attention backbone, which indicates that the
graph bias and the logic objective interact rather than merely add.
 
Third, the synthesized four-level benchmark provides a subject-disjoint
evaluation protocol, a compositional generalisation split, and 487
active constraints derived from its ontology. It constitutes a
controlled resource for evaluating hierarchical intention recognition,
while its designer-defined labels and lack of independently annotated
real intentions remain important limitations.
 
The framework provides a controlled basis for studying predictive
accuracy and ontology-aligned coherence in intention-aware systems; the
present benchmark does not establish operational safety in real robotic
deployment.
Three directions are identified for future extension.
 
\paragraph{Longer sequences and richer hierarchies.}
The current framework processes activity-level episodes of mean 7.04
clips (7.36 on the compositional split). Extension to HLI-level
sequences of 10--20 clips, spanning multiple activities, would test
whether the HR-G transition structure gains discriminative value at
longer horizons, beyond its currently modest contribution relative to
the dominant low-level-intention edge, and would supply the
cross-episode context on which the transitional high-level intentions
are separable. Two further extensions follow from the analysis of the
residual errors: explicit scene-context priors, such as time-of-day or
static environmental anchors, as an additional conditioning input; and
typed message passing between activity, LLI, and HLI nodes layered on
the temporal Transformer rather than substituted for it. Annotation at
five or six semantic levels, including sub-intention and causal reason
layers, would evaluate whether the FOL constraint mechanism generalises
to deeper ontological structures.
 
\paragraph{Anticipatory goal inference.}
The HR-G framework is extended naturally to predict
not only current intentions but also future goals and
anticipated actions.
Coupling the NSGT with a robotic planning framework for
end-to-end anticipatory assistance, where hierarchical
intention predictions directly inform action selection,
is the primary target application.
Evaluation of human trust through a user study in an
assistive scenario is identified as an important
validation step.
 
\paragraph{Online and streaming inference.}
The current framework processes fixed-length episodes
offline.
Real assistive systems require intention inference online,
updating predictions incrementally as new clips arrive.
Adaptation of the NSGT to streaming inference with causal
attention masks and incremental FOL constraint checking is
a direct architectural extension of the current design.

%%═════════════════════════════════════════════════════════════════
\section*{Statements and Declarations}
 
\textbf{Funding.}
This research was carried out within the LISSI Laboratory
(Laboratoire Images, Signaux et Systèmes Intelligents),
University of Paris-Est Créteil (UPEC), Créteil, France.

\textbf{Competing interests.}
The authors declare that they have no competing interests,
financial or non-financial, related to the work submitted
for publication.
 
\textbf{Ethics approval and consent to participate.}
This study does not involve human participants, animals, or
personal data collection.
All experiments are conducted on the publicly available
NTU RGB+D~120 dataset \cite{liu2019ntu}, which was collected
under institutional ethics approval by the original dataset
authors.
No additional ethics approval is required for this work.
 
\textbf{Consent for publication.}
Not applicable.
 
\textbf{Data availability.}
The base dataset used in this study, NTU RGB+D~120, is publicly
available at
\url{https://rose1.ntu.edu.sg/dataset/actionRecognition/}.
The synthesized NTU-Hierarchy benchmark, including episode
manifests, ontology mappings, and FOL rule matrices, will be
made available upon acceptance of this manuscript.
 
\textbf{Materials availability.}
Not applicable.
 
\textbf{Code availability.}
The implementation code for the NSGT framework, including
training scripts, model definitions, and evaluation notebooks,
will be made available upon acceptance of this manuscript.
 
\textbf{Author contributions.}
F.~Soleimani: Conceptualization, Methodology, Software,
Formal analysis, Investigation, Writing (original draft).
G.~Khodabandelou: Supervision, Conceptualization,
Writing (review and editing).
A.~Chibani: Supervision, Writing (review and editing).
Y.~Amirat: Supervision, Writing (review and editing).

\begin{appendices}
 
\section{High-Level Intention Error Structure Across Model Families}
\label{secA1}
 
The high-level intention confusion matrices of three representative
models are compared to determine whether the residual errors at this
level reflect architectural limitations or the intrinsic structure of
the ontology. The three models are the external relational-graph
baseline (R-GCN), the strongest sequential baseline (Seq-Transformer),
and the proposed architecture without the logic objective (NSGT). The
logic-free variant is used so that the comparison isolates the
graph-structured architecture itself. The effect of the logic objective
is a cross-level property, reported through the hierarchy consistency
rate in the main text, and does not appear in a single-level confusion
matrix.
 
\begin{figure}[H]
\centering
\begin{subfigure}[t]{0.32\textwidth}
  \includegraphics[width=\textwidth]{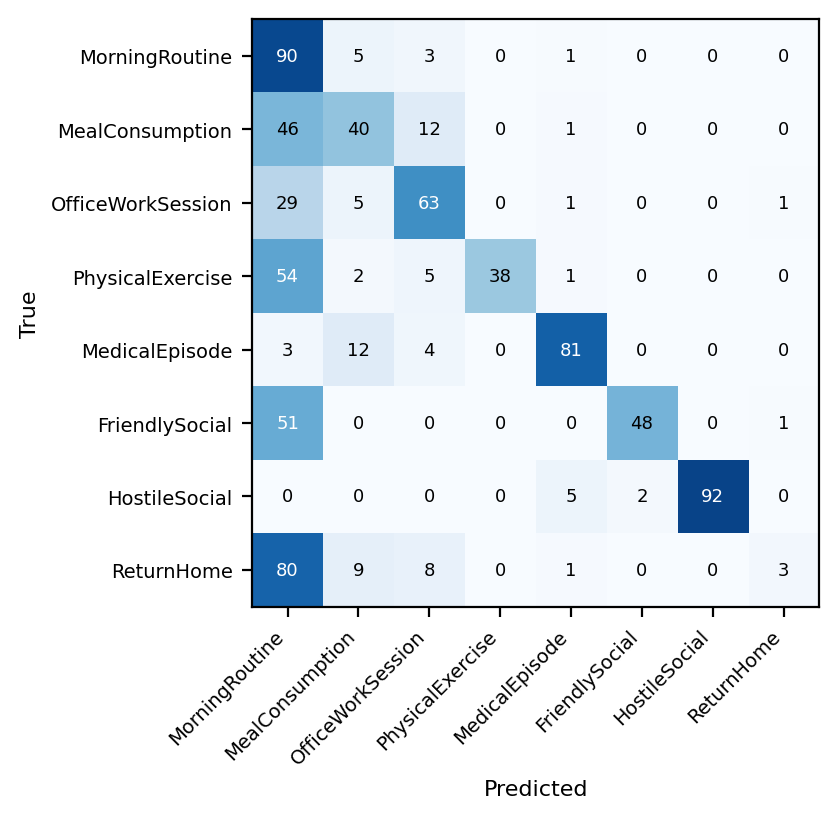}
  \caption{R-GCN}
\end{subfigure}
\hfill
\begin{subfigure}[t]{0.32\textwidth}
  \includegraphics[width=\textwidth]{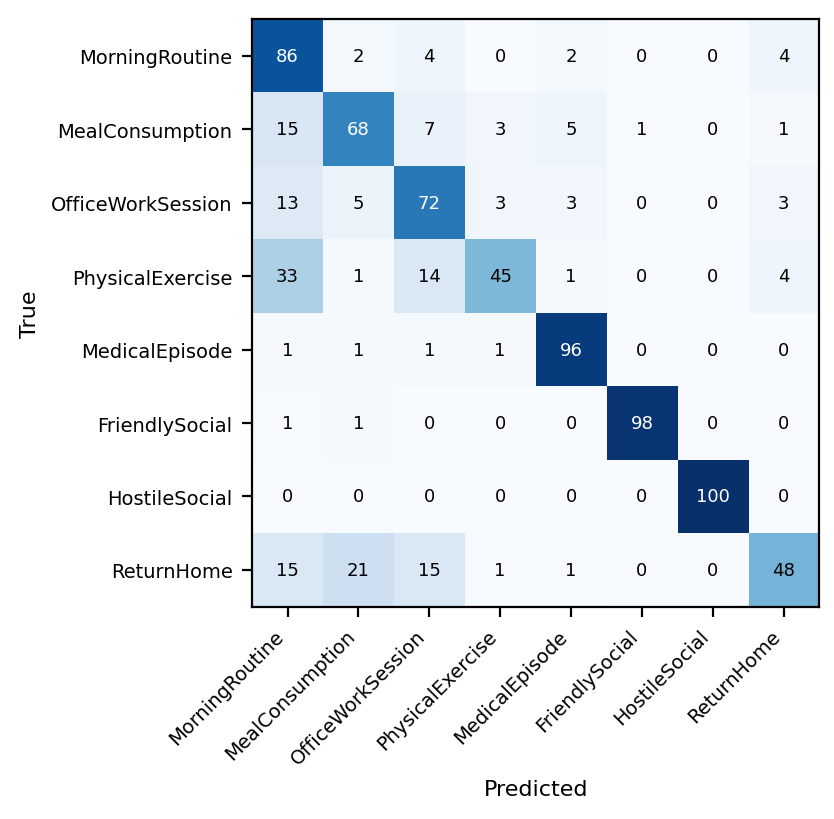}
  \caption{Seq-Transformer}
\end{subfigure}
\hfill
\begin{subfigure}[t]{0.32\textwidth}
  \includegraphics[width=\textwidth]{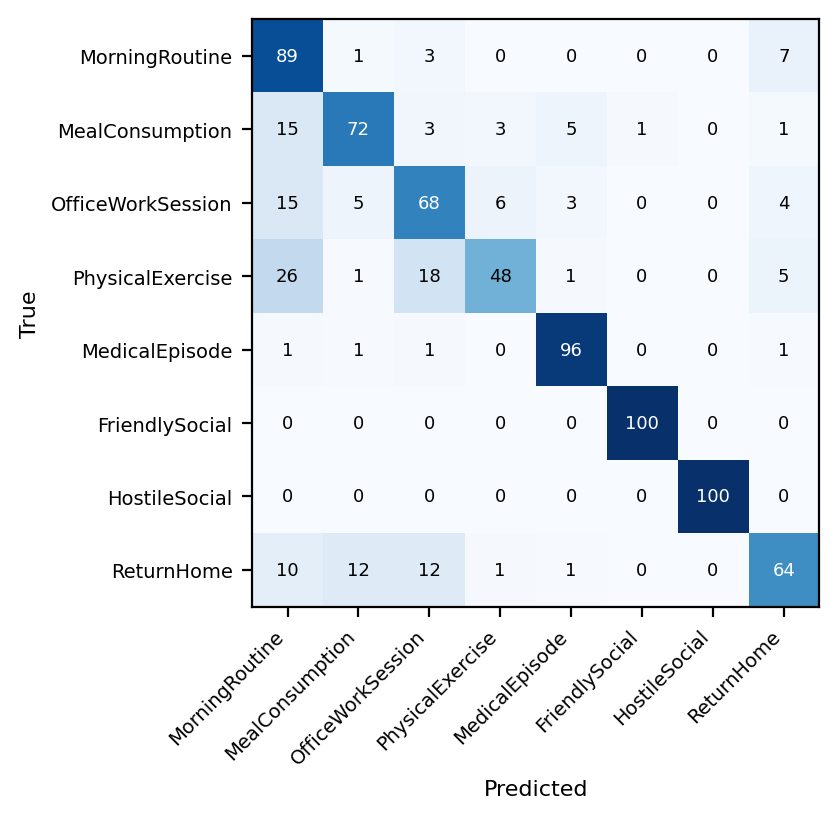}
  \caption{NSGT (no FOL)}
\end{subfigure}
\caption{
  Row-normalised high-level-intention confusion matrices (test set,
  values in \%, seed 42) for three model families. Error magnitude
  decreases from the relational-graph baseline to the sequential
  baseline to the graph-structured architecture, but the more
  informative difference is in the structure of the errors, which is
  analysed below.
}
\label{fig:cm_hli_family}
\end{figure}
 
\paragraph{The confused classes share the same transitional intentions.}
The errors are not distributed uniformly. They concentrate on four
high-level intentions, morning routine, office work, physical exercise,
and return-home, while three others, medical episode, friendly social,
and hostile social, are recognised almost perfectly by the proposed
model (96.5\%, 99.7\%, and 100.0\% respectively). This partition follows
directly from the ontology. The four confused classes each draw three
or four of their constituent low-level intentions from a small set of
transitional intentions, walking transit, seated rest, phone check, and
drink break, and each of these four low-level intentions is a
constituent of four distinct high-level intentions. Walking transit,
for example, is a valid sub-intention of morning routine, office work,
physical exercise, and return-home simultaneously. The three classes
that are recognised cleanly contain none of these shared transitional
intentions, and are instead composed of distinctive sub-intentions with
a single natural parent. The errors therefore fall precisely on the
high-level intentions that are ontologically many-to-one at the
intention level, and avoid the classes that are structurally
unambiguous.
 
\paragraph{Kinematic symmetry compounds the ontological overlap.}
The overlap is not only symbolic. The shared transitional intentions
correspond to motor patterns that are close to identical in the
skeleton stream: return-home and morning routine both consist of
transitional behaviour such as walking, handling outerwear, and picking
up or putting down objects, and both draw on the walking-transit and
phone-check intentions. Separating such episodes therefore depends on
static scene and object context rather than on body pose, which is
precisely the dependence quantified by the modality ablation in
Section~\ref{sec:modality_results}, where these classes collapse under
skeleton-only input and are restored by the appearance stream. The
residual confusions among the four transitional classes are thus the
joint consequence of a many-to-one ontology and an input whose
kinematic component is symmetric across them.
 
\paragraph{The baselines fail on this overlap in a characteristic way.}
The relational-graph baseline performs static message passing over the
ontological edges and does not model the temporal order of the atomic
actions within an episode. When two episodes present overlapping atomic
actions, as the transitional classes do, the baseline has no
fine-grained temporal evidence with which to break the tie, and it
collapses toward the majority attractor: 80\% of return-home episodes
are routed to morning routine, and over half of physical-exercise
episodes are routed to morning routine as well, even though these pairs
share only their transitional intentions and are otherwise distinct.
The result is a diffuse error pattern that spills across ontologically
distant classes. The Seq-Transformer removes most of this diffuse
error, since it models the action sequence, but with no relational
prior it still recognises physical exercise and return-home below 50\%,
concentrating its residual error on the same transitional classes.
 
\paragraph{The residual limitation of the proposed model is a
representational boundary, not an architectural one.}
The graph-structured architecture resolves the intention-defined and
socially defined boundaries cleanly, recognising hostile and friendly
social episodes at or above 99.7\%, which shows that where the
distinction is carried by the intention structure the model recovers it
in full. Its remaining errors lie entirely on the transitional classes
identified above, whose separation depends on evidence that the input
representation does not encode: static scene context such as the
environment, the objects in view, and time-of-day anchors. The model
receives pooled clip-level video and skeleton features and no explicit
scene-context prior, so where two high-level intentions are legitimately
reachable from the same physical sub-actions, and differ only in
context, the representation is genuinely underdetermined. This is a
property of the input and the ontology, not a failure of the
architecture to represent hierarchy: the same model resolves every
boundary that the ontology makes distinctive. It also points to a
concrete direction for improvement, namely the injection of explicit
scene-context priors, which is discussed in the main text as future
work.
 
\end{appendices}

 
 
 
%\end{appendices}
 
 
\bibliography{sn-bibliography}% common bib file
\end{document}